\documentclass{article}

\usepackage{microtype}
\usepackage{graphicx}
\usepackage{subfigure}
\usepackage{booktabs} 

\usepackage{hyperref}

\usepackage[accepted]{icml2023}

\usepackage{amsmath}
\usepackage{amssymb}
\usepackage{mathtools}
\usepackage{amsthm}
\usepackage{graphbox}
\usepackage{xcolor}
\usepackage[most]{tcolorbox}
\usepackage{soulpos}
\usepackage{multirow}

\usepackage[capitalize,noabbrev]{cleveref}

\theoremstyle{plain}

\theoremstyle{definition}

\theoremstyle{remark}

\NewDocumentCommand\emojimaple{}{{\includegraphics[scale=0.05]{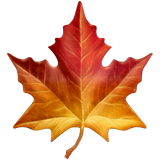}\hspace{1mm}}}

\NewDocumentCommand\emojilight{}{{\includegraphics[scale=0.05]{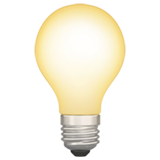}}}
\NewDocumentCommand\emojipinkheart{}{{\includegraphics[scale=0.05]{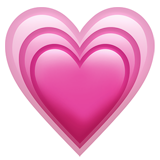}}}

\usepackage[textsize=tiny]{todonotes}

\icmltitlerunning{Gradient-Based Discrete Optimization for Prompt Tuning and Discovery\hfill\thepage}
\begin{document}

\twocolumn[
\icmltitle{\centering Hard Prompts Made Easy: \\
Gradient-Based Discrete Optimization for Prompt Tuning and Discovery}

\hypersetup{
    colorlinks=true,
    pdftitle={Hard Prompts Made Easy: Gradient-Based Discrete Optimization for Prompt Tuning and Discovery},
    pdfpagemode=FullScreen,
    }

%



\icmlsetsymbol{equal}{*}

\begin{icmlauthorlist}
\icmlauthor{Yuxin Wen}{equal,$^1$}
\icmlauthor{Neel Jain}{equal,$^1$}
\icmlauthor{John Kirchenbauer}{$^1$}
\icmlauthor{Micah Goldblum}{$^2$}
\icmlauthor{Jonas Geiping}{$^1$}
\icmlauthor{Tom Goldstein}{$^1$}
\end{icmlauthorlist}

\begin{center}
\textbf{$^1$University of Maryland, $^2$New York University} \\
\{ywen, njain17, jkirchen, jgeiping, tomg\}@umd.edu, goldblum@nyu.edu
\end{center}



\vskip 0.3in
]



\printAffiliationsAndNotice{\icmlEqualContribution} 

\begin{abstract}
The strength of modern generative models lies in their ability to be controlled through text-based prompts. 
Typical  ``hard'' prompts are made from interpretable words and tokens, and must be hand-crafted by humans.  There are also ``soft'' prompts, which consist of continuous feature vectors.  These can be discovered using powerful optimization methods, but they cannot be easily interpreted,  re-used across models, or plugged into a text-based interface.

We describe an approach to robustly optimize hard text prompts through efficient gradient-based optimization. Our approach automatically generates hard text-based prompts for both text-to-image and text-to-text applications. In the text-to-image setting, the method creates hard prompts for diffusion models, allowing API users to easily generate, discover, and mix and match image concepts without prior knowledge on how to prompt the model. In the text-to-text setting, we show that hard prompts can be automatically discovered that are effective in tuning LMs for classification. 

\end{abstract}

\section{Introduction}
\label{intro}

Prompt engineering is the art of creating instructions to guide generative models.  It is the key to unlocking the power of large models for both image generation and language tasks. As it stands today, prompt engineering methods can be coarsely divided into two camps.  First, there are {\em hard} prompting methods, which use hand-crafted sequences of interpretable tokens to elicit model behaviors. Hard prompt discovery is a specialized alchemy, with many good prompts being discovered by trial and error, or sheer intuition.   Then there are {\em soft} prompts, which consist of continuous-valued language embeddings that do not correspond to any human-readable tokens.  Soft prompt discovery is a mathematical science;  gradient-based optimizers and large curated datasets are used to generate highly performant prompts for specialized tasks. 

\begin{figure}[t]
    \centering
    \vspace{-.25cm}
    \includegraphics[width=\columnwidth]{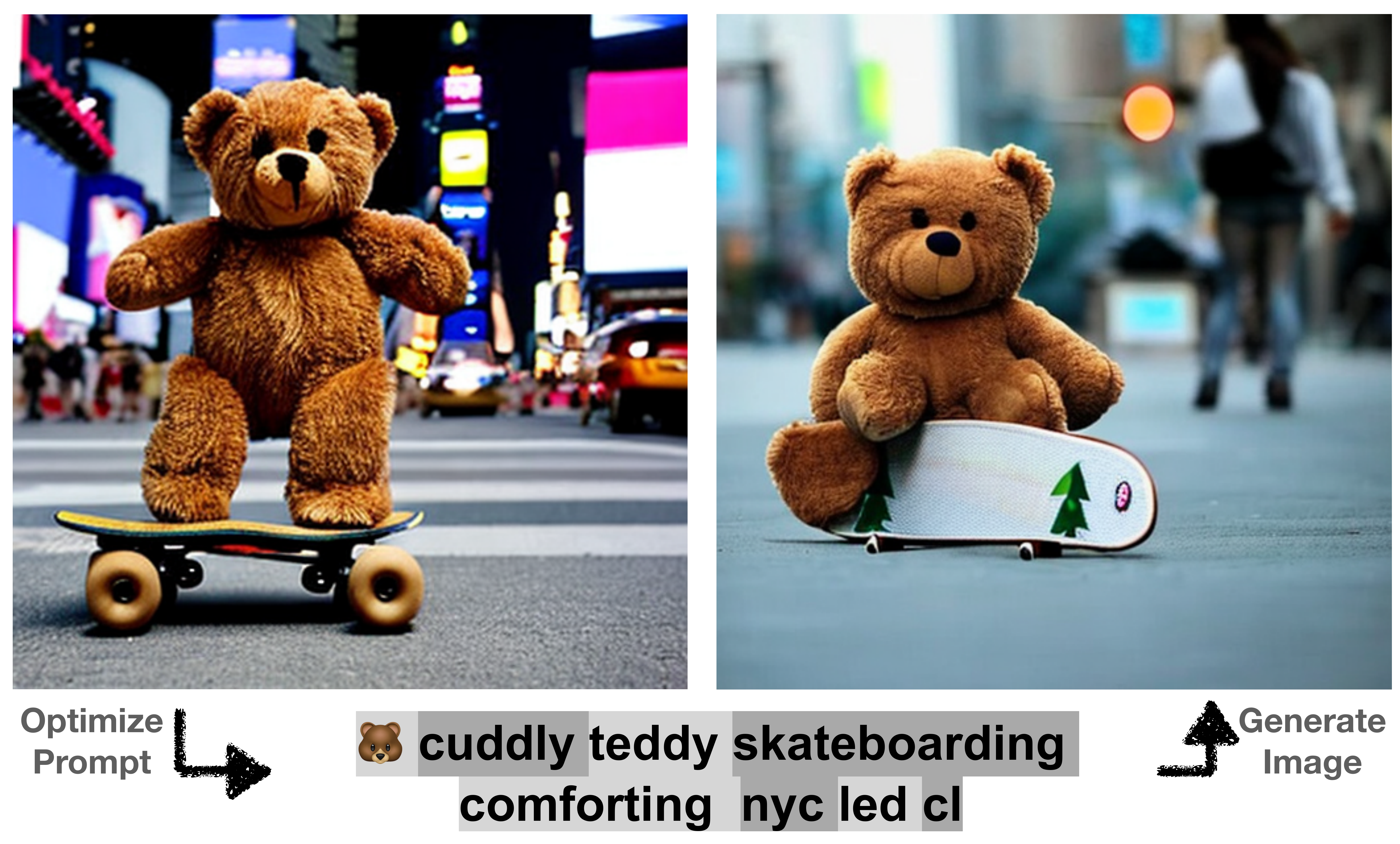}
    \includegraphics[width=\columnwidth]{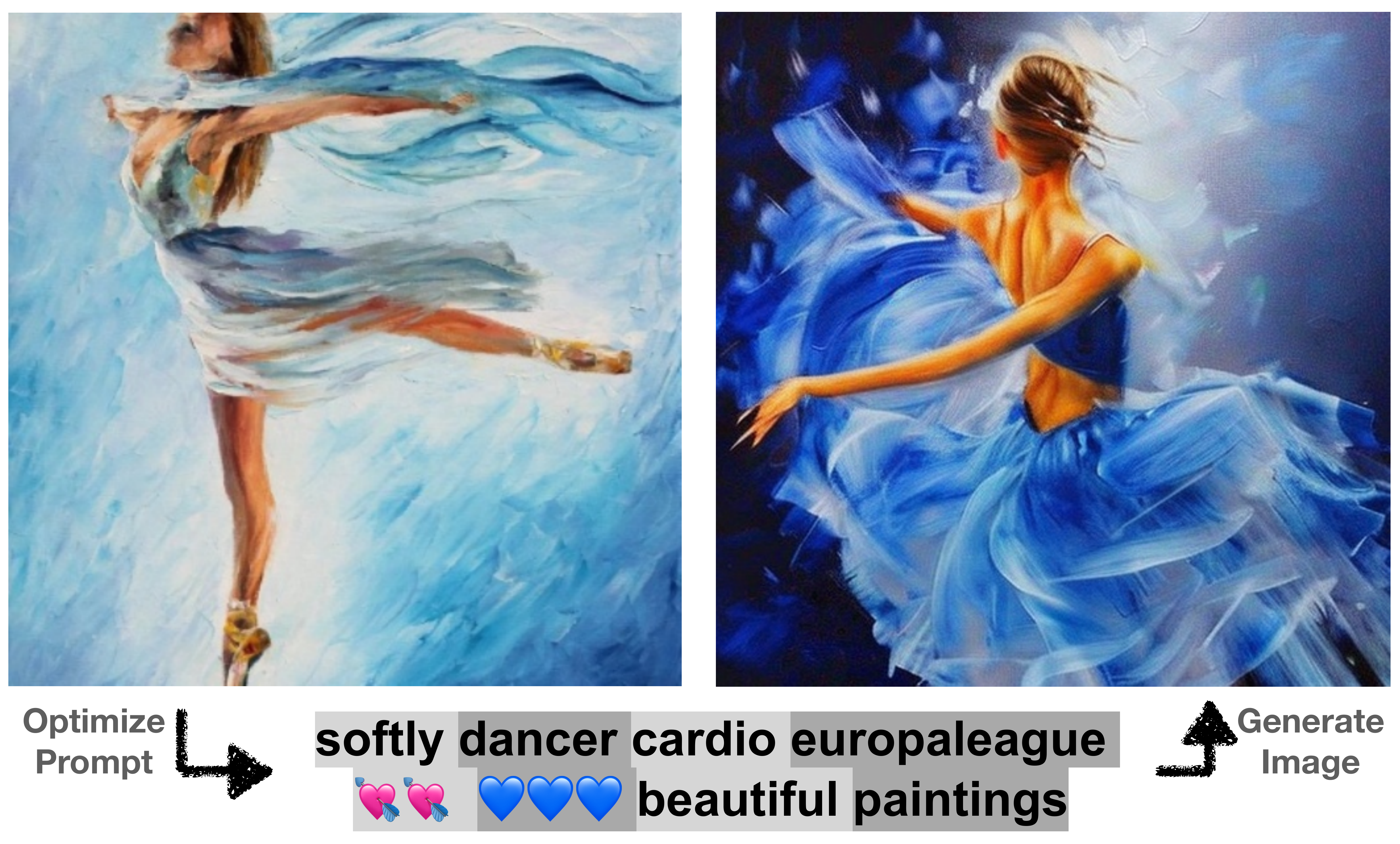}
    \vspace{-.45cm}
    \caption{Two examples of hard prompt discovery through optimization. Given an image (left), a discrete text prompt is discovered using CLIP and used to prompt Stable Diffusion, generating new images (right). Two shades of gray are used to show the token boundaries in the recovered prompt.}
    \label{fig:teaser_figure}
    \vspace{-.5cm}
\end{figure}

Despite the difficulty of engineering hard prompts, they have their advantages. Hard prompts and the tricks they exploit can be mixed, matched, and mutated to perform a range of different tasks, while soft prompts are highly specialized.  Hard prompts are portable; they can be discovered using one model and then deployed on another.  This portability is impossible with soft prompts due to differences in embedding dimension and representation space between models.  Finally, hard prompts can be used when only API access to a model is available and it is not possible to control the embeddings of inputs.

This work explores the use of efficient gradient methods to optimize and learn discrete text, with an emphasis on applications to prompt engineering. In doing so, we unlock the ability to learn hard prompts via optimization.  Learned hard prompts combine the ease and automation of soft prompts with the portability, flexibility, and simplicity of hard prompts.  Our primary contributions are summarized as follows:
\begin{itemize}
\item  We propose a simple scheme for learning hard prompts using continuous optimization.  The scheme builds on existing gradient reprojection schemes for optimizing text, and adapts lessons learned from the large-scale discrete optimization literature for quantized networks.
\item We show that this optimization method can be used to learn hard prompts for image generation, giving us a general tool to create prompts that elicit specific image styles, objects, and appearances.  The learned prompts perform competitively with highly specialized prompt generation tools, despite using far fewer tokens and containing no hand-crafted components.    
\item We also show that our learned hard prompts perform well on language classification tasks, out-performing other text optimization schemes. The learned prompts transfer well across networks, and this transfer is enhanced when they are regularized with fluency constraints to improve interpretability.
\end{itemize}
In addition to capturing the quantifiable benefits of learned prompts, the proposed schemes 
can be used to facilitate {\em prompt exploration and discovery}, as optimization often recovers words and tokens that are simultaneously highly interpretable and also highly non-obvious.

\section{Related Works}

\textbf{Prompting in Language Models.}
\citet{brown_language_2020_2} was one of the first to demonstrate the
power of prompting for task adaption of pre-trained language models.
This ``instruction tuning'' paradigm has since become a standard way to increase the ability of large models to follow complex, task-specific instructions \citep{sanh2022multitask, chung2022scaling}.
However, automatically finding suitable sets of text prompts, i.e. hard prompts, for these purposes remains an open challenge.
\citet{lester_power_2021} simplified the ``prefix tuning'' technique presented in \citet{li2021prefix} to establish the procedure referred to as standard \emph{soft} ``prompt-tuning'' where they optimize sequences of continuous-valued embeddings prepended to the real embeddings of the input tokens. However, subsequent work by \citet{khashabi-etal-2022-prompt} showed that the sequences of embeddings produced by this technique could map to token sequences with limited semantic scrutability. To address these limitations, in this work we construct a method for hybridizing the continuous soft-prompt optimization with hard vocabulary constraints, resulting in task-specific, interpretable tokens.

\looseness -1 \textbf{Discrete Optimization for Language.}\label{sec:discr-lang-opt}
AutoPrompt \citep{shin-etal-2020-autoprompt} was one of the first discrete prompt optimization frameworks for transformer language models and subsequent approaches have included a gradient-free phrase editing method \citep{prasad2022grips}, an embedding optimization approach based on Langevin dynamics \citep{shi2022toward} and a reinforcement learning approach \citep{Deng2022RLPromptOD}. 

We consider two gradient-based methods as baselines: \textit{FluentPrompt} and \textit{AutoPrompt} \citep{shi2022toward, shin-etal-2020-autoprompt}. \textit{AutoPrompt}, which utilizes \textit{HotFlip} proposed by
\citet{ebrahimi-etal-2018-hotflip}, greedily chooses the optimal token for each location in the prompt utilizing the gradient to find a selection of good candidates. However, \textit{AutoPrompt} can become expensive very quickly. For each gradient step, the method requires an evaluation of each candidate at each location in the prompt, adding numerous additional forward passes. To avoid the additional forward passes, we originally considered $\textit{AutoPrompt}_{k=1}$ with and without an added fluency constraint, but found that $\textit{AutoPrompt}_{\text{SGD}}$ with a fluency constraint outperformed its counterparts as seen in \cref{fig:AutoPromptSGDvsK1}, and thus we use SGD version of \textit{AutoPrompt} as our other baseline similar to \citet{shi2022toward}. \textit{FluentPrompt} differs from \textit{AutoPrompt} by utilizing Langevin dynamics \citep{kumar2022gradient} to optimize the prompt embeddings, as well as adding a fluency penalty.

\looseness -1 For the baselines discussed above, at the end of every update step, the optimized prompt embeddings are projected onto their nearest neighbor embeddings to ensure that optimization is performed on the discrete set of natural language tokens. 
However, if the nearest neighbors are far away from the embeddings and the learning rate is not tuned properly, the embeddings may become stagnant, which can require extensive hyperparameter tuning as demonstrated in \cref{fig:FluentPrompt_LRsweep}. 
The cost of such a constraint is a loss of flexibility in the solutions the optimization can find.
On the other hand, 
while soft prompts are not as limited in this way,
just clamping a well-trained soft prompt to the nearest discrete prompt
strongly degrades performance as observed in \citet{khashabi-etal-2022-prompt}.

\looseness -1 \textbf{Prompt Discovery from Images.}
The process of extracting rich information from images and conveying it through natural language texts is known as \emph{image captioning}. \citet{zhang2021vinvl}, \citet{hu2022scaling}, and \citet{li2022blip} achieve this goal by training large captioning models on image-text pairs. However, these captions are often generic and may not accurately reflect new or unseen objects. In \citet{gal2022image}, the authors propose a method that utilizes a soft prompt to optimize a text-guided diffusion model, allowing for the generation of similar visual concepts to those in the original image. In this case, though the final soft prompt is effective, optimization through a diffusion model is very expensive, and the prompts are neither interpretable nor portable.

\textbf{Discrete Optimization.}
Discrete optimizers
have long been used to train neural networks with quantized (e.g. binary) weights.  In that context, the approach of re-projecting between gradient steps is known as {\em stochastic rounding}.  However, it is known that this approach lacks the convergence guarantees of continuous optimization \cite{li2017training}.   Over the last decade, stochastic rounding has been replaced by newer optimizers that maintain a continuous, rather than discrete, representation of the weights \citep{courbariaux2015binaryconnect}.  These optimizers consistently result in higher accuracy \citep{rastegari2016xnor,courbariaux2016binarized} and avoid local minima \citep{li2017training}.

We take inspiration from these lessons learned in the binary networks community and adapt them to refine and simplify discrete optimizers for language.

\section{Methodology}

\looseness -1 \textbf{Learning Hard Prompts.}
We now present our effective and easy-to-use technique for discrete prompt optimization. The process requires the following inputs: a frozen model, $\theta$, a sequence of learnable embeddings, $\mathbf{P}=[\mathbf{e_i}, ... \mathbf{e_M}], \mathbf{e_i} \in \mathbb{R}^{d} $, where $M$ is the number of ``tokens'' worth of vectors to optimize, and $d$ is the dimension of the embeddings. Additionally, we employ an objective function $\mathcal{L}$. The discreteness of the token space is realized using a projection function, $\text{Proj}_{\mathbf{E}}$, that takes the individual embedding vectors $\mathbf{e_i}$ in the prompt and projects them to their nearest neighbor in the embedding matrix $E^{|V|\times d}$ where $|V|$ is the vocabulary size of the model, and we denote the result of this operation as  $\mathbf{P'}=\text{Proj}_{\mathbf{E}}(\mathbf{P}):=[\text{Proj}_{\mathbf{E}}(\mathbf{e_i}), ... \text{Proj}_{\mathbf{E}}(\mathbf{e_M})]$. 
Additionally, we define a broadcast function, 
$\mathcal{B}: \mathbb{R}^{(M\times d)} \to \mathbb{R}^{(M\times d\times b)}$ that repeats the current prompt embeddings ($\mathbf{P}$) in the batch dimension $b$ times.

Formally, to learn a hard prompt, we minimize the following risk by measuring the performance of $\mathbf{P}$ on the task data: $R(\mathbf{P'}) = \mathbb{E}_{D}(\mathcal{L(\theta(\mathcal{B}(\mathbf{P}, \mathbf{X})), \mathbf{Y})})$.

\begin{algorithm}[t]
   \caption{Hard \textbf{P}rompts made \textbf{E}a\textbf{Z}y: PEZ Algorithm}
   \label{alg:pez}
\begin{algorithmic}[Pez]
\STATE \textbf{Input:} Model $\theta$, vocabulary embedding $\mathbf{E}^{|V|}$, projection function $\text{Proj}$, broadcast function $\mathcal{B}$, optimization steps $T$, learning rate $\gamma$, Dataset $D$
\STATE {\color{gray} Sampled from real embeddings:}
\STATE $\mathbf{P}=[\mathbf{e_i}, ... \mathbf{e_M}] \sim \mathbf{E}^{|V|}$ 
\FOR{$1, ..., T$}
    \STATE Retrieve current mini-batch $(X, Y) \subseteq D$.
    \STATE {\color{gray} Forward Projection:}
    \STATE $\mathbf{P'}=\text{Proj}_{\mathbf{E}}(\mathbf{P})$
    \STATE {\color{gray}Calculate the gradient w.r.t. the \textit{projected} embedding:}
    \STATE $g = \nabla_{\mathbf{P'}} \mathcal{L_{\text{task}}}(\mathcal{B}(\mathbf{P'}, X_i), Y_i, \theta)$
    \STATE {\color{gray} Apply the gradient on the \textit{continuous} embedding:}
    \STATE $\mathbf{P} = \mathbf{P} - \gamma g$
\ENDFOR
\STATE {\color{gray} Final Projection:}
\STATE $\mathbf{P} = \text{Proj}_{\mathbf{E}}[\mathbf{P}]$
\STATE \textbf{return} $\mathbf{P}$
\end{algorithmic}
\end{algorithm}
\textbf{Our Method.}
\looseness -1 We propose a simple but efficient gradient-based discrete optimization algorithm that combines the advantages of the baseline discrete optimization methods and soft prompt optimization.
The steps of our scheme, which we call PEZ, are concretely defined in \cref{alg:pez}.
The method
maintains continuous iterates, which in our applications corresponds to a soft prompt.
During each forward pass, we first project the current embeddings $\mathbf{P}$ onto the nearest neighbor $\mathbf{P}'$ before calculating the gradient. Then, using the gradient of the discrete vectors, $\mathbf{P}'$, we update the continuous/soft iterate, $\mathbf{P}$. 

\begin{figure*}[t]
    \centering
    \begin{tabular}{c@{\hspace{20pt}}c@{\hspace{2pt}}c@{\hspace{2pt}}c@{\hspace{2pt}}c}
        Target Image & \multicolumn{4}{c}{Generated Image with Learned Hard Prompt} 
    \\
        \includegraphics[align=c, scale=0.16]{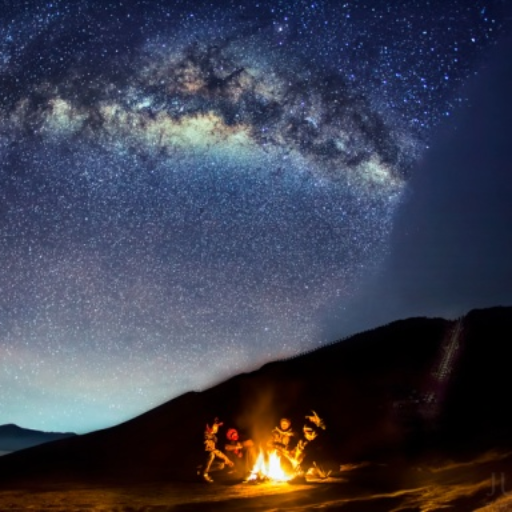} & 
        \includegraphics[align=c, scale=0.16]{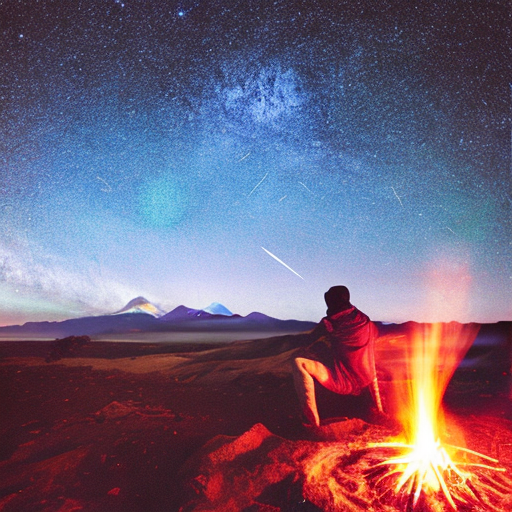} &
        \includegraphics[align=c, scale=0.16]{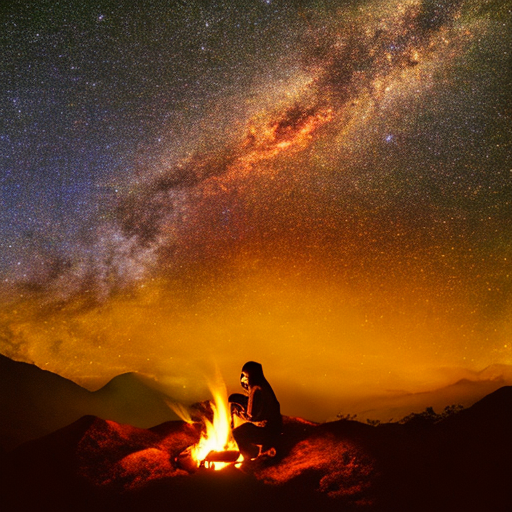} &
        \includegraphics[align=c, scale=0.16]{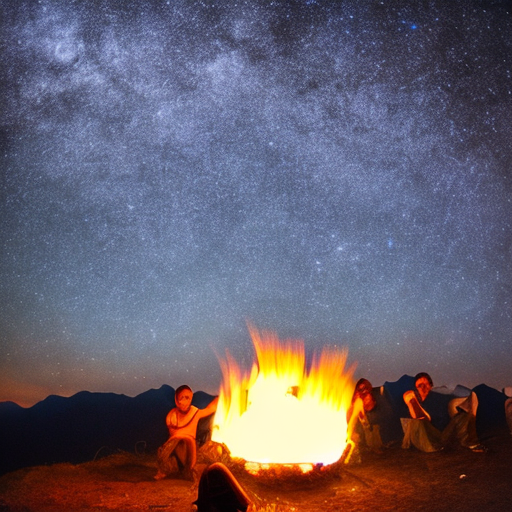} &
        \includegraphics[align=c, scale=0.16]{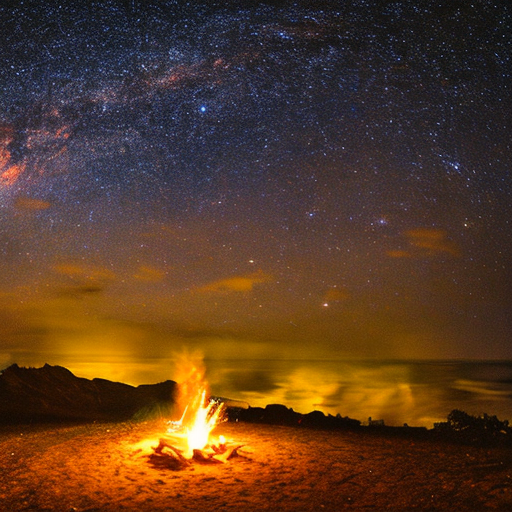}
    \\ [3pt]
        &
        \multicolumn{4}{c}{\begin{tabular}[c]{c}
            \includegraphics[align=c, scale=0.16]{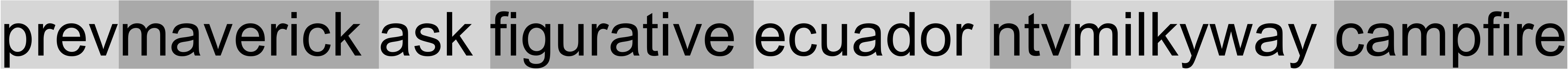}
        \end{tabular}}
    \\ [3pt]
        \includegraphics[align=c, scale=0.16]{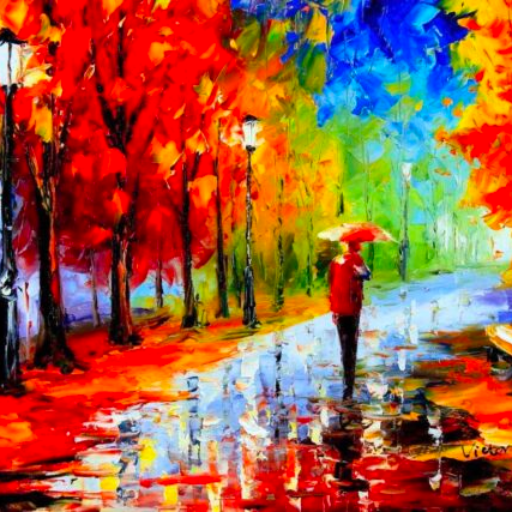} & 
        \includegraphics[align=c, scale=0.16]{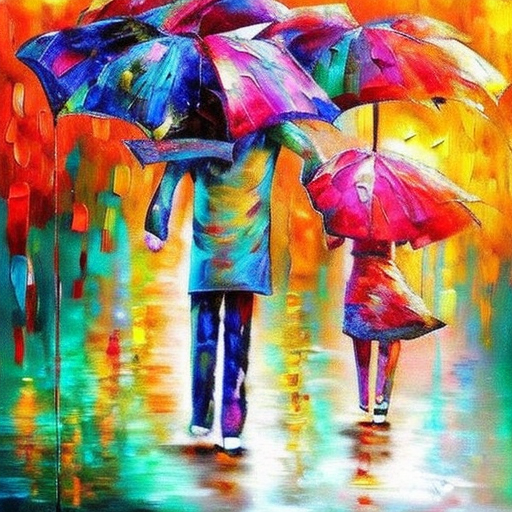} &
        \includegraphics[align=c, scale=0.16]{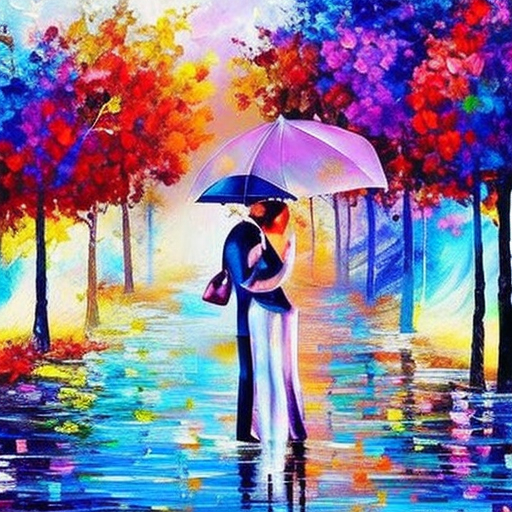} &
        \includegraphics[align=c, scale=0.16]{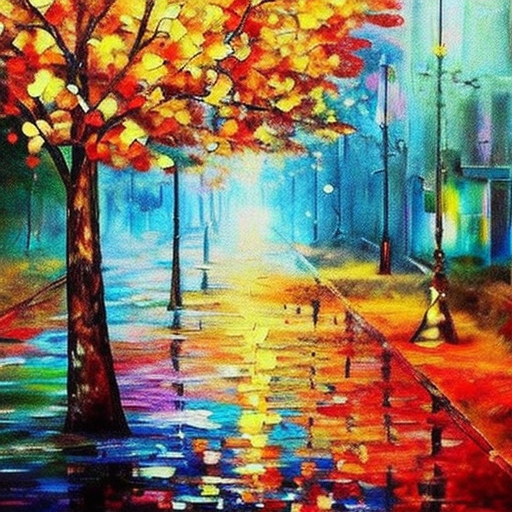} &
        \includegraphics[align=c, scale=0.16]{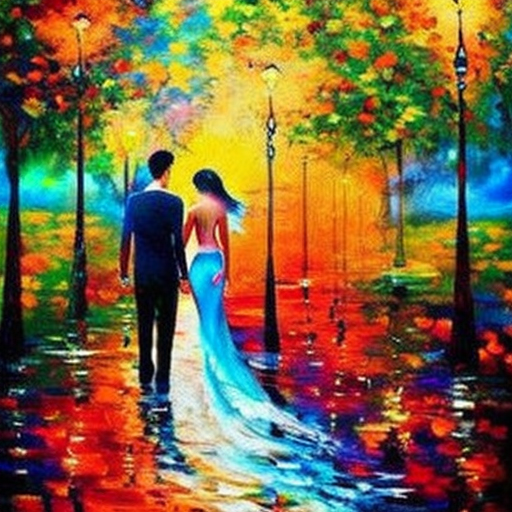}
    \\ [3pt]
        & 
        \multicolumn{4}{c}{\begin{tabular}[c]{c}
            \includegraphics[align=c, scale=0.16]{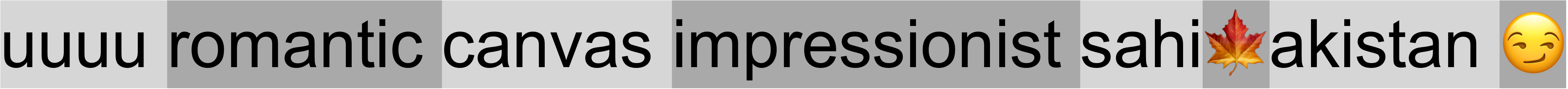}
        \end{tabular}}
    \\ [3pt]
        \includegraphics[align=c, scale=0.16]{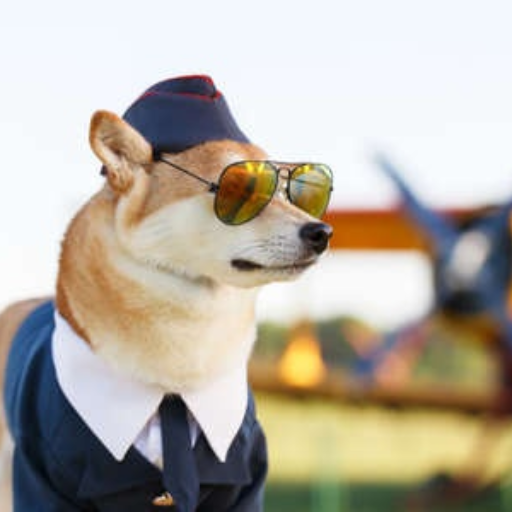} &
        \includegraphics[align=c, scale=0.16]{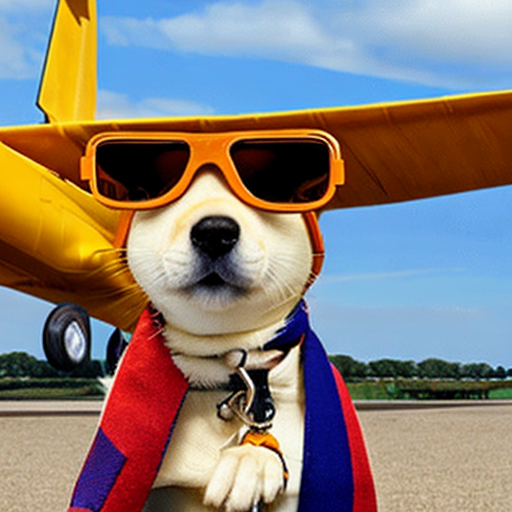} &
        \includegraphics[align=c, scale=0.16]{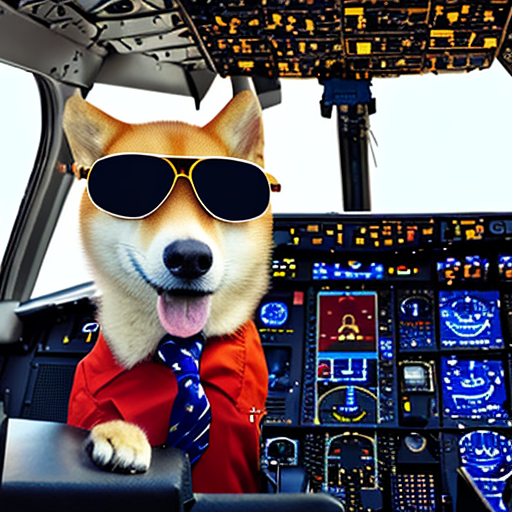} &
        \includegraphics[align=c, scale=0.16]{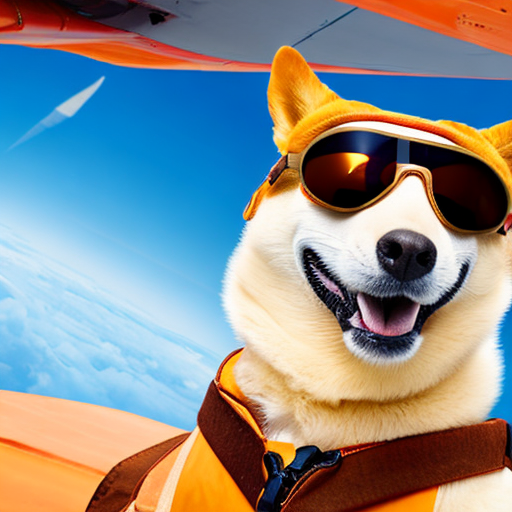} &
        \includegraphics[align=c, scale=0.16]{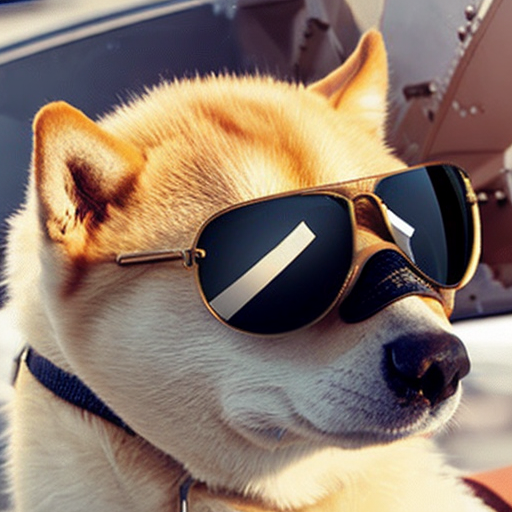}
    \\ [3pt]
        &
        \multicolumn{4}{c}{\begin{tabular}[c]{c}
            \includegraphics[align=c, scale=0.16]{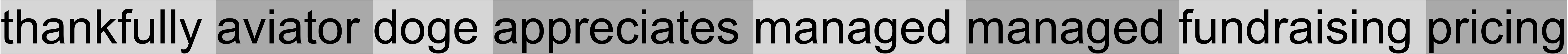}
        \end{tabular}}
    \\ [3pt]
        \includegraphics[align=c, scale=0.16]{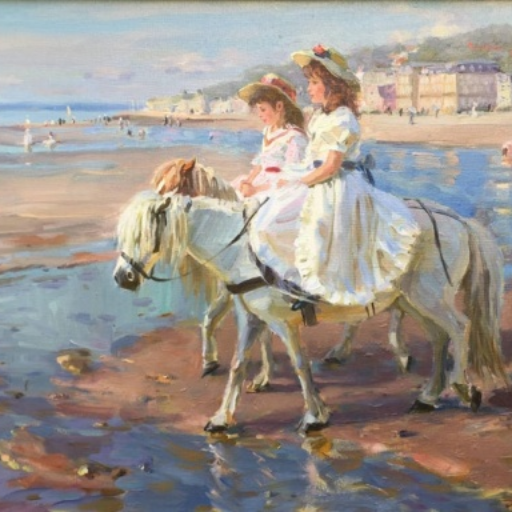} & 
        \includegraphics[align=c, scale=0.16]{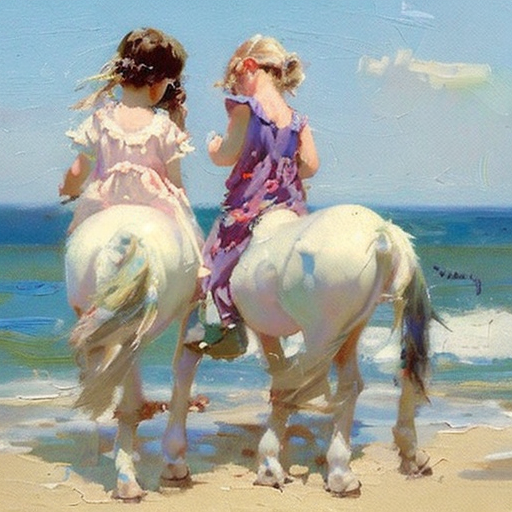} &
        \includegraphics[align=c, scale=0.16]{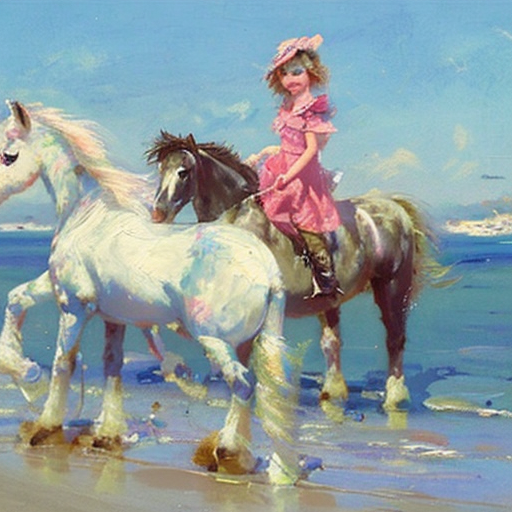} &
        \includegraphics[align=c, scale=0.16]{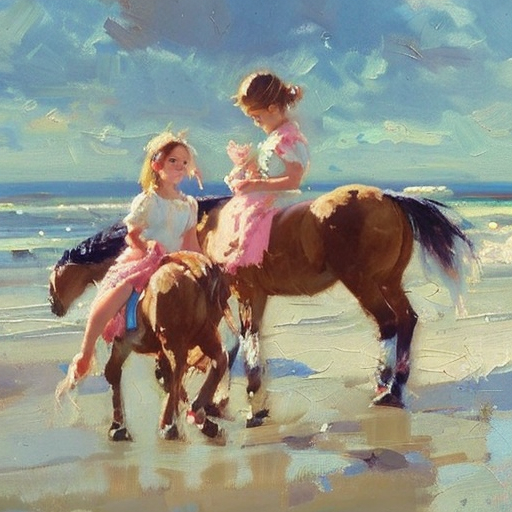} &
        \includegraphics[align=c, scale=0.16]{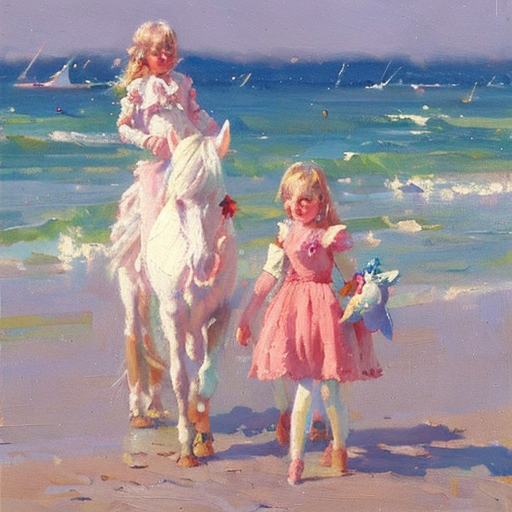}
    \\ [3pt]
        &
        \multicolumn{4}{c}{\begin{tabular}[c]{c}
            \includegraphics[align=c, scale=0.16]{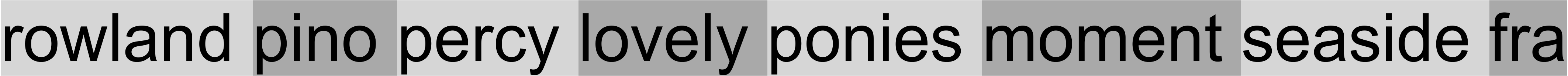}
        \end{tabular}}
    \\
    \end{tabular}
    \caption{Generations using learned hard prompts on four different target images. For a given target image (left), a discrete text prompt is discovered using CLIP and used to prompt Stable Diffusion, generating new images (right). Two shades of gray are used to show the token boundaries in the recovered prompt.}
    \label{fig:clip-main}
\end{figure*}

\section{Prompt Inversion with CLIP}
Our method for learning hard prompts is perfectly suited to multimodal vision-language models. 
With these models, like CLIP \citep{radford2021learning}, we can use PEZ to discover captions which describe one or more target images. 
In turn, these discovered captions can be deployed as prompts for image generation applications. Since most text-guided diffusion models utilize pre-trained text encoders, such as the CLIP text encoder, and freeze them during training, we can discover prompts using these pre-trained text encoders that are directly relevant for downstream diffusion models.  For instance, we can optimize a caption which describes an image and use this caption as a prompt for a diffusion model to generate other images with the same content.

Since the CLIP model has its own image encoder,
we can leverage it as a loss function to drive our PEZ method. This way we are optimizing prompts only for their cosine similarity to the CLIP image encoder, and avoiding gradient calculations on the full diffusion model altogether. 

Formally, given a text encoder function $f$ and an image encoder function $g$, we optimize the hard prompt embedding $\mathbf{P}$ corresponding to a target image $x$ by minimizing the following objective: $\mathcal{L}(\mathbf{P},x) = 1 - \mathcal{S}(f(\mathbf{P}),g(x))$, where $\mathcal{S}$ is the cosine similarity between two vectors.

\subsection{Experimental Setting}
We conduct experiments on four datasets with diverse distributions: LAION \citep{schuhmann2022laion}, MS COCO \citep{lin2014microsoft}, Celeb-A \citep{liu2015deep}, and Lexica.art \citep{santana_gustavostastable-diffusion-prompts_2022}. LAION comprises over $5$ billion diverse images scraped from the internet, including photos and paintings.
MS COCO mainly contains real-life photographs with multiple common objects, whereas Celeb-A consists of celebrity portraits. Lexica.art is a set of AI-generated paintings with their prompts.

We measure the quality of the prompt via image similarity between original (target) image, and an image generated using the learned hard prompt. To do so, we use a larger reference CLIP model, OpenCLIP-ViT/G, that was not used during optimization and serves as a neutral metric for semantic similarity between the images.

We choose Stable Diffusion-v2 \citep{rombach2022high} as our generative model, and the open-source CLIP model, OpenCLIP-ViT/H \citep{cherti2022reproducible} for crafting the prompt, as both share the same text encoder. During the prompt optimization process, we use a generic learning rate of $0.1$ and run $3000$ optimization steps using the AdamW optimizer \citep{loshchilov2017decoupled}. For Stable Diffusion-v2, we set the guidance scale to $9$ and the number of inference steps to $25$. For each dataset, we randomly sample $100$ data points and average CLIP scores over $5$ runs with different random seeds.

\begin{table*}[t]
\centering
\small
\caption{Quantitative evaluation of learned hard prompts. We report the CLIP score between the original images and the images generated by the hard prompts.  A high score indicates that generated and target images contain similar semantic content.}
\label{table:clip-main}
\begin{tabular}{ccccccc}
\toprule
                 & \#Tokens   & Requirement         & LAION & MS COCO & Celeb-A & Lexica.art \\ \midrule
PEZ (Ours)             & $8$   & CLIP                & $0.697$ & $0.674$   & $0.602$   & $0.711$      \\
CLIP Interrogator & $\sim77$  & CLIP + Bank + BLIP & $0.707$ & $0.690$   & $0.558$   & $0.762$      \\ \midrule
CLIP Interrogator without BLIP  & $\sim77$ & CLIP + Bank        & $0.677$ & $0.674$   & $0.572$   & $0.737$      \\
PEZ (Ours) + Bank      & $8$ & CLIP + Bank        & $0.702$ & $0.689$   & $0.629$   & $0.740$      \\ \midrule
CLIP Interrogator    & $8$ & CLIP + Bank + BLIP & $0.539$ & $0.575$   & $0.360$   & $0.532$      \\
CLIP Interrogator   & $16$ & CLIP + Bank + BLIP & $0.650$ & $0.650$    & $0.491$   & $0.671$      \\
CLIP Interrogator  & $32$ & CLIP + Bank + BLIP & $0.694$ & $0.663$   & $0.540$  & $0.730$      \\ \midrule
Soft Prompt   & $8$ & CLIP & $0.408$ & $0.420$   & $0.451$   & $0.554$      \\ \bottomrule
\end{tabular}
\end{table*}
\begin{figure*}[t]
    \centering
    \begin{tabular}{c@{\hspace{20pt}}c@{\hspace{2pt}}c@{\hspace{2pt}}c@{\hspace{2pt}}c}
        Target Style & \multicolumn{4}{c}{Learned Hard Prompt + keywords} 
    \\
        \includegraphics[align=c, scale=0.16]{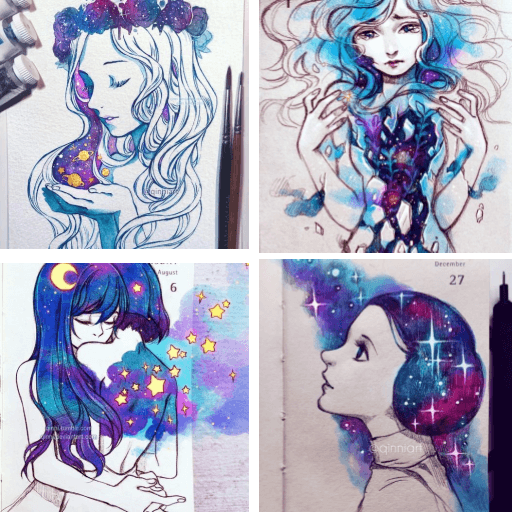} & 
        \includegraphics[align=c, scale=0.16]{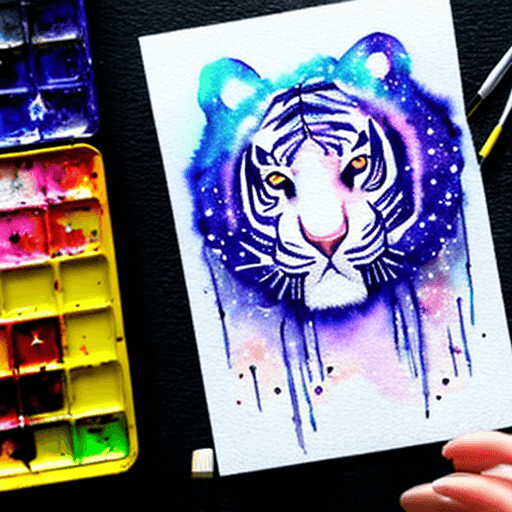} &
        \includegraphics[align=c, scale=0.16]{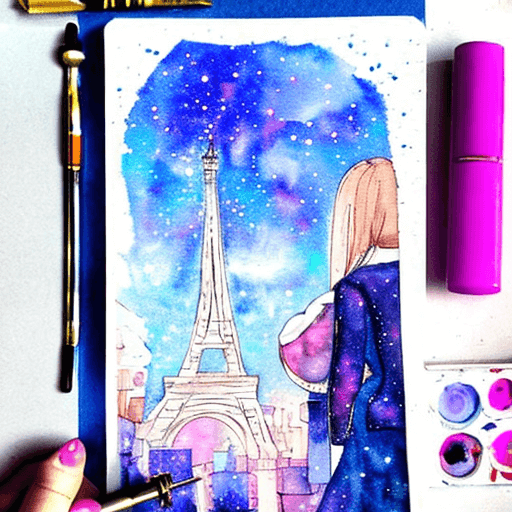} &
        \includegraphics[align=c, scale=0.16]{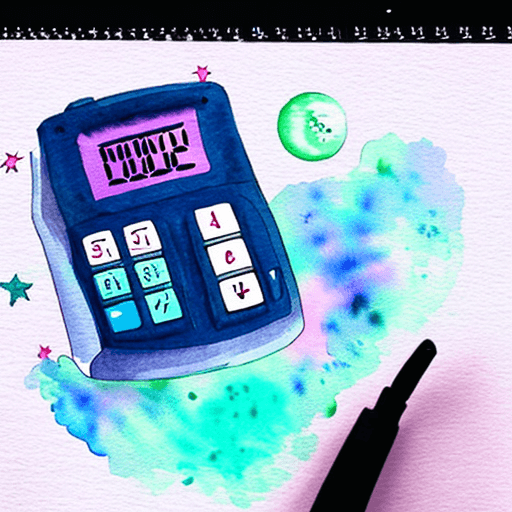} &
        \includegraphics[align=c, scale=0.16]{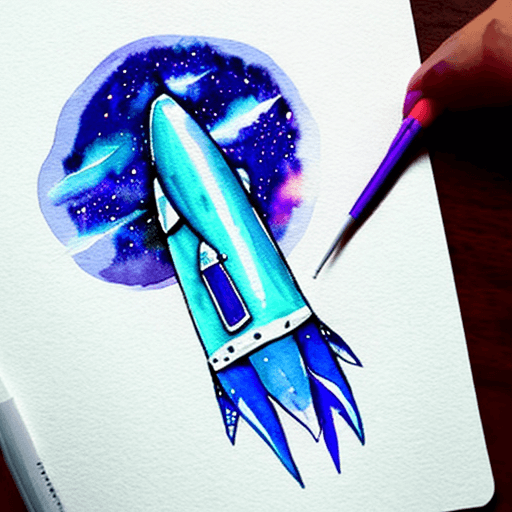}
    \\ [40pt]
        \includegraphics[align=c, scale=0.16]{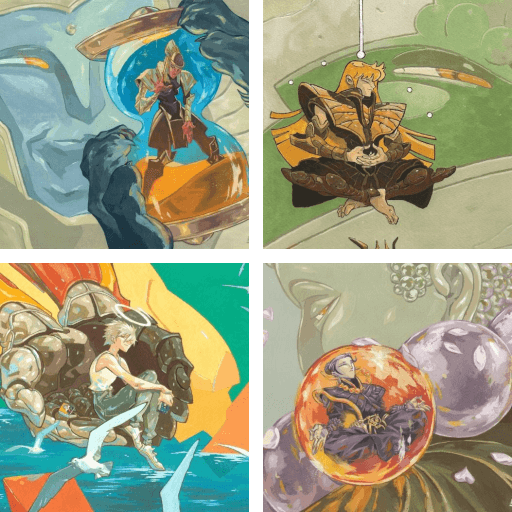} & 
        \includegraphics[align=c, scale=0.16]{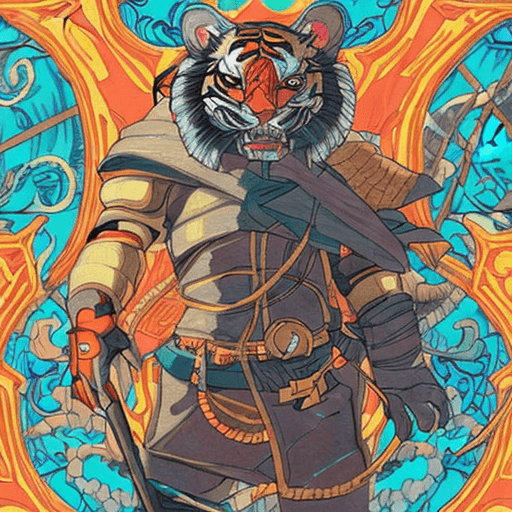} &
        \includegraphics[align=c, scale=0.16]{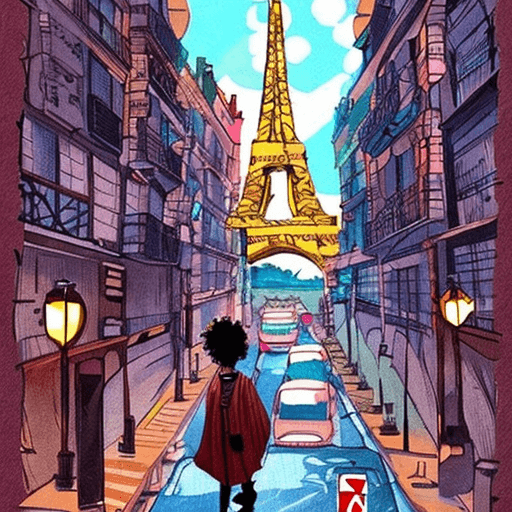} &
        \includegraphics[align=c, scale=0.16]{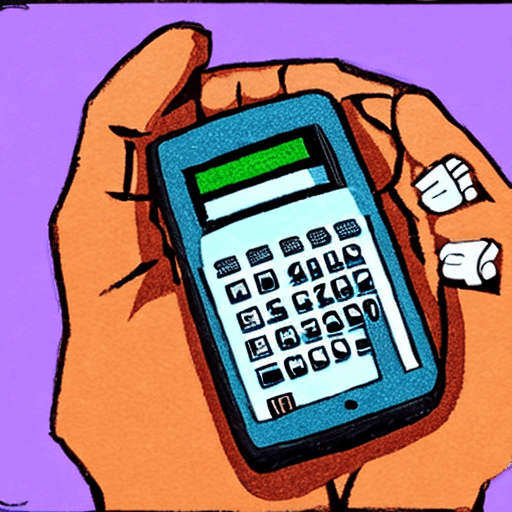} &
        \includegraphics[align=c, scale=0.16]{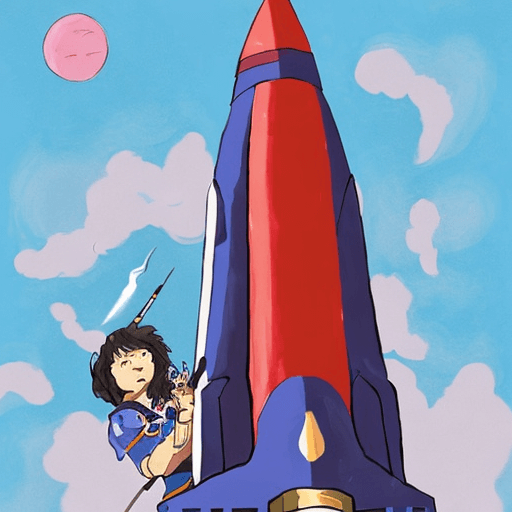}
    \\
        &
        A tiger &
        Paris &
        A calculator &
        A rocket
    \end{tabular}
    \caption{Learned hard prompts for style transfer. Given several sample images with the same style, we can extract the style with a hard prompt and transfer it to other objects or scenes. Detailed templates and hard prompts can be found in \cref{app:clip}. Sample images credits: \href{https://www.deviantart.com/qinni}{Qinni} and \href{https://www.deviantart.com/facundo-lopez}{facundo-lopez}.}
    \label{fig:clip-style}
\end{figure*}

\looseness -1 A natural baseline for hard prompt discovery with CLIP is the \textit{CLIP Interrogator}\footnote{\small \url{https://github.com/pharmapsychotic/clip-interrogator}}. To generate a descriptive hard prompt, this tool first uses a pre-trained captioning model, BLIP \citep{li2022blip} to create a caption of the target image. Then, top-$k$ keywords from a pre-collected bank of keywords are appended to the caption based on CLIP scores between the keywords and the target image. These keywords were collected from various sources, including 5,265 artist names like ``Van Gogh'' and 100,970 phrases from prompt engineering, resulting in a diverse set.
We find this keyword bank to contain most of the phrases from the Lexica.art dataset. \textit{CLIP Interrogator} then greedily samples keywords until the prompt reaches CLIP's token length limit of $77$.

\begin{figure*}[!hbt]
    \centering
    \begin{tabular}{c@{\hspace{2pt}}c@{\hspace{2pt}}c@{\hspace{2pt}}c@{\hspace{2pt}}c@{\hspace{2pt}}c@{\hspace{2pt}}c}
        \multicolumn{3}{c}{Separate Generation} & & \multicolumn{3}{c}{Concatenated Generation} 
    \\
        \includegraphics[align=c, scale=0.16]{assets/samples/set4/s4_g4.png} & 
        $+$ &
        \includegraphics[align=c, scale=0.16]{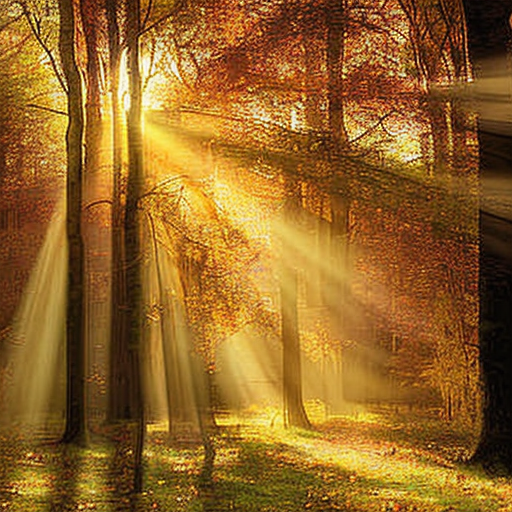} &
        $=$ &
        \includegraphics[align=c, scale=0.16]{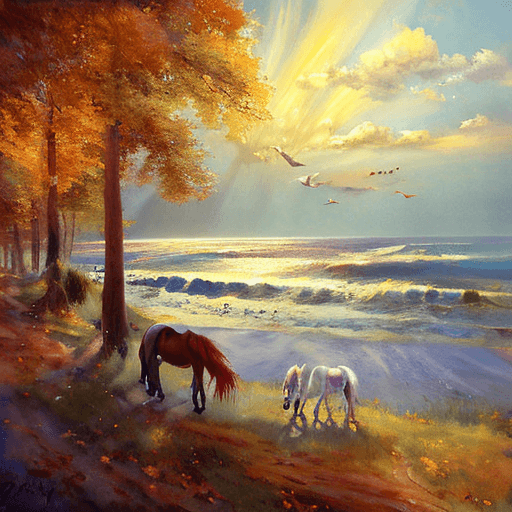} &
        \includegraphics[align=c, scale=0.16]{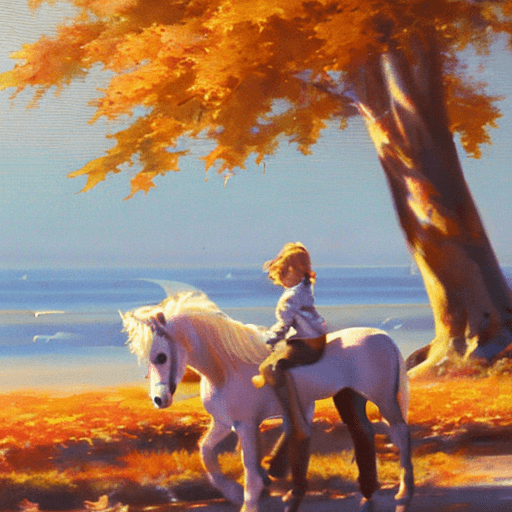} &
        \includegraphics[align=c, scale=0.16]{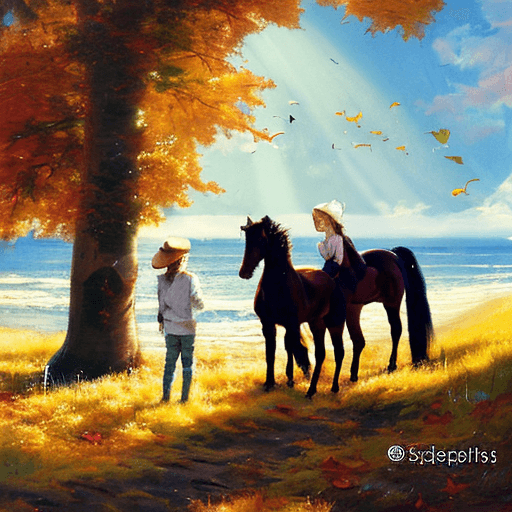}
    \\ [3pt]
        \multicolumn{7}{c}{\begin{tabular}[c]{c}
            \includegraphics[align=c, scale=0.16]{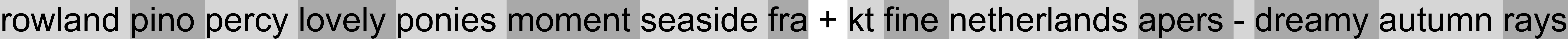}
        \end{tabular}}
    \\ [3pt]
        \includegraphics[align=c, scale=0.16]{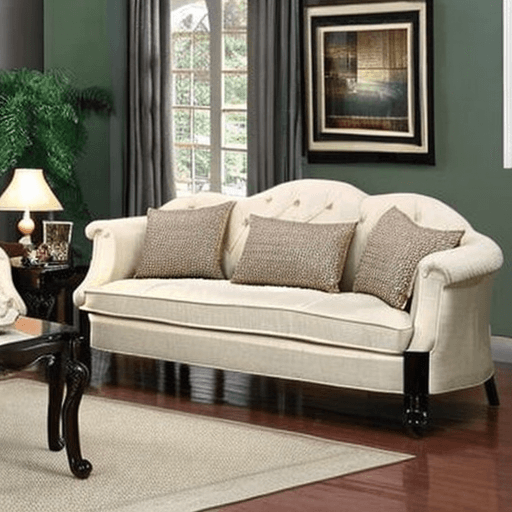} & 
        $+$ &
        \includegraphics[align=c, scale=0.16]{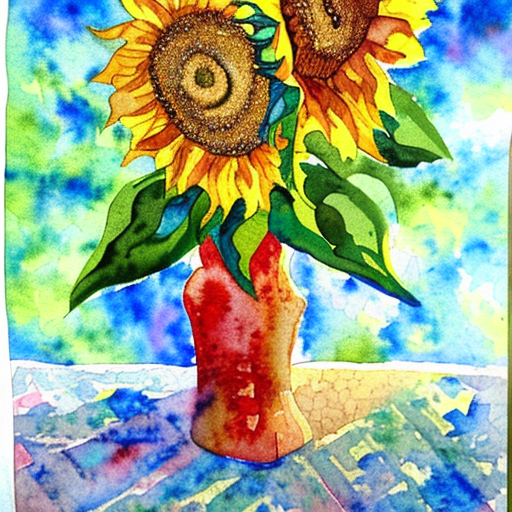} &
        $=$ &
        \includegraphics[align=c, scale=0.16]{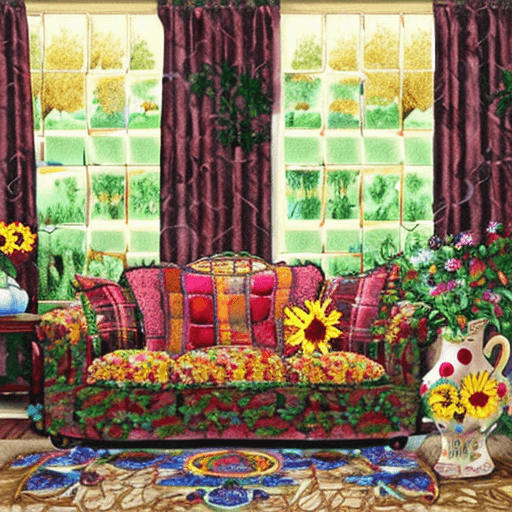} &
        \includegraphics[align=c, scale=0.16]{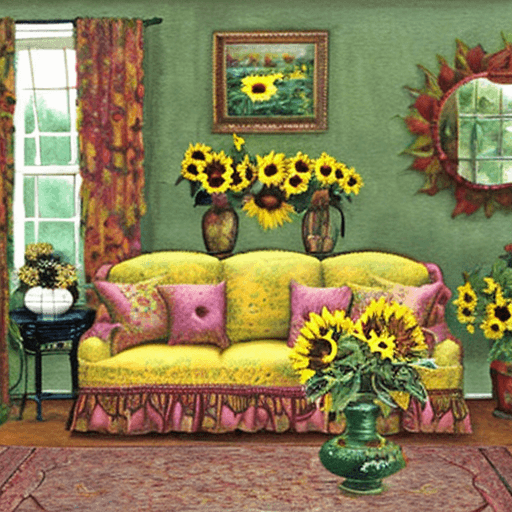} &
        \includegraphics[align=c, scale=0.16]{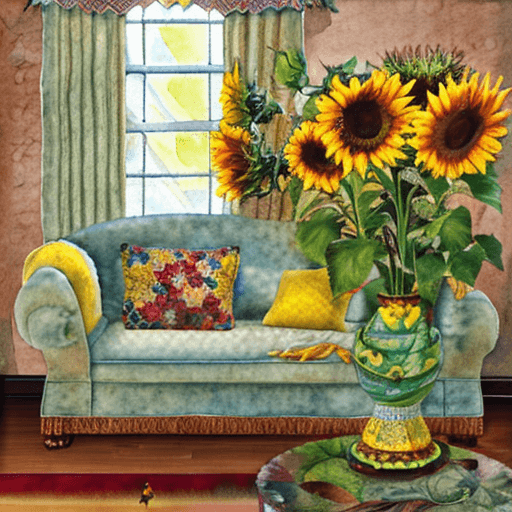}
    \\ [3pt]
        \multicolumn{7}{c}{\begin{tabular}[c]{c}
            \includegraphics[align=c, scale=0.16]{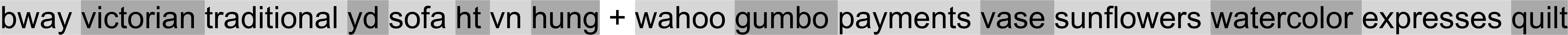}
        \end{tabular}}
    \end{tabular}
    \caption{Concatenated learned hard prompts. We show the hard prompts learned on two unrelated images can be concatenated to fuse the semantic concepts in them.}
    \label{fig:clip-merge}
    \vspace{-.25cm}
\end{figure*}
\subsection{Results}

We show example hard prompts learned using our method and corresponding generations in \cref{fig:clip-main}. The generated images clearly show that the prompts effectively capture the semantic features of the target images. Further, the generations are highly similar to the original images as measured by CLIP score and under visual inspection. Additionally, the hard prompts do not overfit to the original target image and produce a diverse set of generated images given different random seeds.

Prompts are human readable, containing a mix of real words and gibberish (non-word token sequences). However, the valid words that are included in the prompts provide a significant amount of information about the image. For example, in the first row, we can see the words ``milkyway'' and ``campfire,'' which are the two main elements in the target image. Interestingly, the optimized prompts may also include emojis, like \emojimaple 
present in the second row. \emojimaple represents the trees on the side and also the color theme of the image.
The optimization process
seems to choose these emojis to include useful information while keeping the prompt concise.

Further, we present quantitative evaluations in \cref{table:clip-main}. Our method performs consistently across all four datasets and outperforms other gradient-based optimization baselines (full table can be found in \cref{table:app-clip-main}).
Notably, we can achieve similar performance to \textit{CLIP Interrogator}, which has the highest CLIP score on LAION, MS COCO, Lexica.art, but not Celeb-A (The keyword bank in \textit{CLIP Interrogator} does not include many words related to real human faces). However, \textit{CLIP Interrogator} uses a large curated prompt dataset, the image captioning model BLIP, and a large number of tokens (as many as $77$), while our proposed method only uses the CLIP model for prompt discovery and $8$ tokens in total demonstrating its simultaneous simplicity and strength. 

We ablate each of these differences. To do so, we include the keyword bank in our optimization method and only allow projections onto tokens from the keyword bank.
Overall, we find that when adding this constraint to our model, and disabling BLIP to compare both methods on equal footing, we recover most of the quantitative difference between the methods on LAION and Lexica.art. Additionally, reducing the token length for the \textit{CLIP Interrogator}, leads to a sharp drop in performance, again, particularly when normalizing by comparing both approaches at equal token lengths of $8$.
\looseness -1 We note that even though Stable Diffusion and CLIP share the same text encoder, \textit{soft prompts do not transfer well} compared to all hard prompt methods in our evaluation.

\begin{figure}
    \centering
    \includegraphics[width=0.48\textwidth,trim={0cm 0.3cm 0cm 0cm},clip]{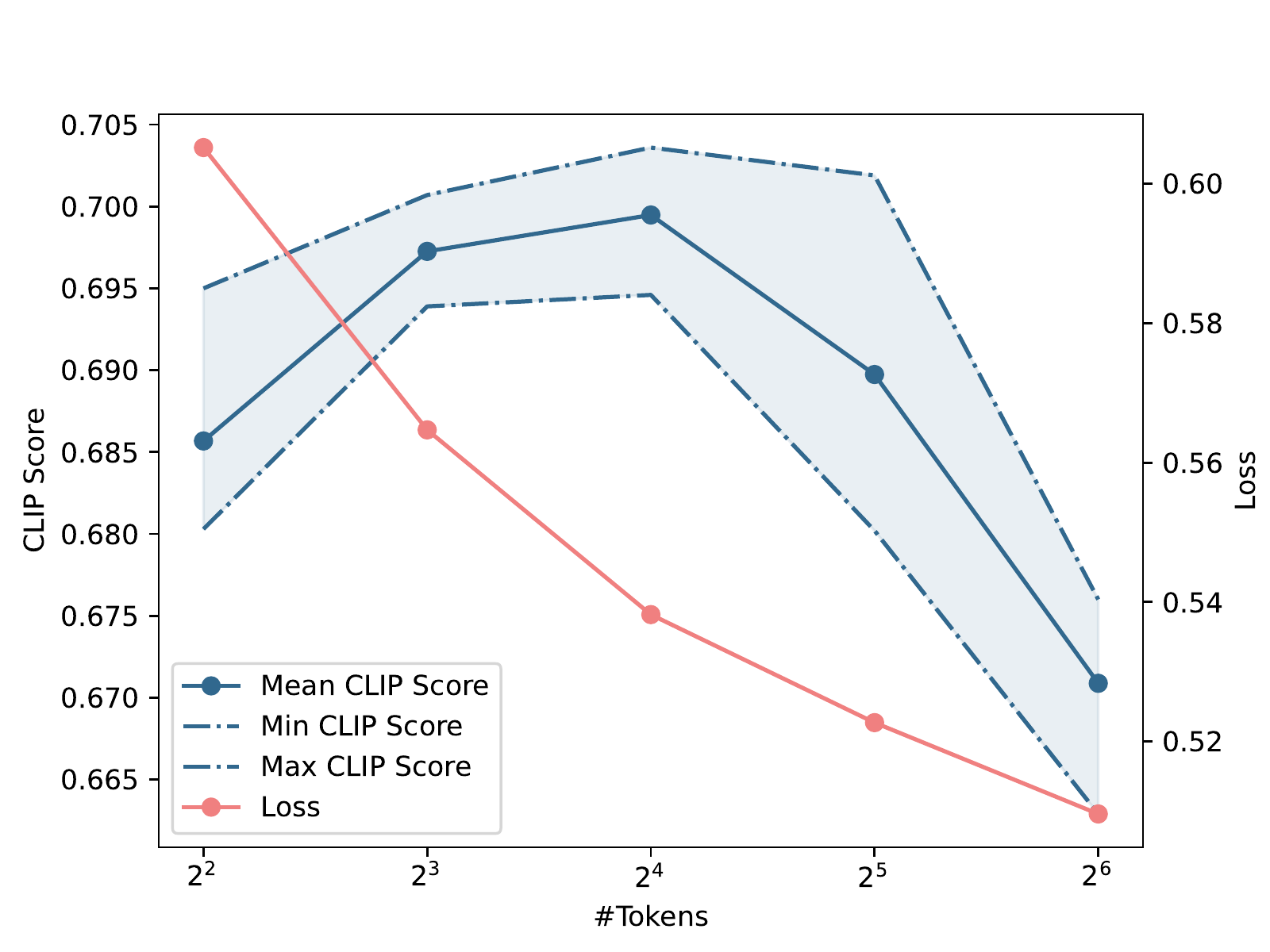}
    \vspace{-.45cm}
    \caption{Ablation on prompt length, showing both train loss on the clip image encoder and validation CLIP score to generated Stable Diffusion images as prompt length increases.}
    \label{fig:prompt-len}
    \vspace{-.5cm}
\end{figure}

\textbf{Prompt Length.} We further ablate the optimal number of tokens. In \cref{fig:prompt-len}, we find that longer prompts do not necessarily produce better results when generating with Stable Diffusion, even though they strictly reduce loss on the CLIP image encoder. Long prompts thus overfit and are less transferable, and we empirically find a length of $16$ to result in the most generalizable performance.

\begin{figure*}[!ht]
    \centering
    \begin{tabular}{c@{\hspace{20pt}}c@{\hspace{2pt}}c@{\hspace{2pt}}c@{\hspace{2pt}}c}
        Target Prompt & \multicolumn{4}{c}{Learned Hard Prompts} 
    \\
        \includegraphics[align=c, scale=0.16]{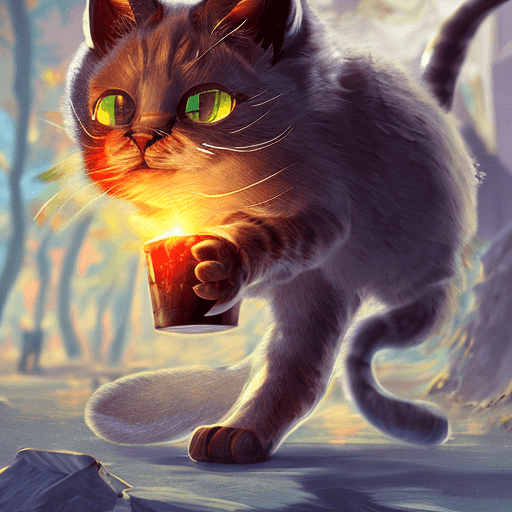} & 
        \includegraphics[align=c, scale=0.16]{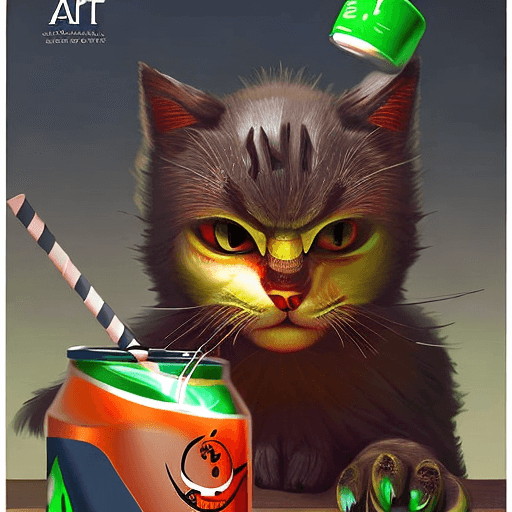} &
        \includegraphics[align=c, scale=0.16]{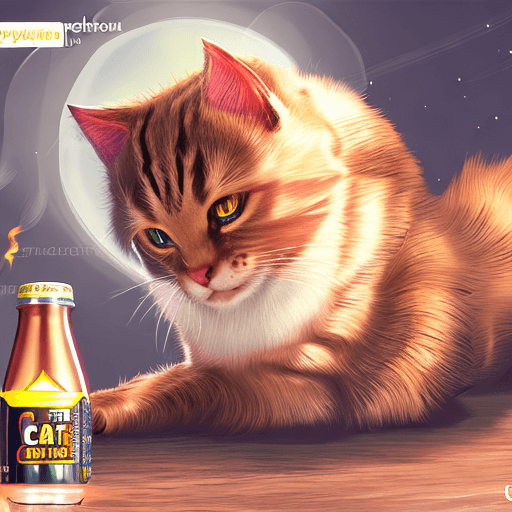} &
        \includegraphics[align=c, scale=0.16]{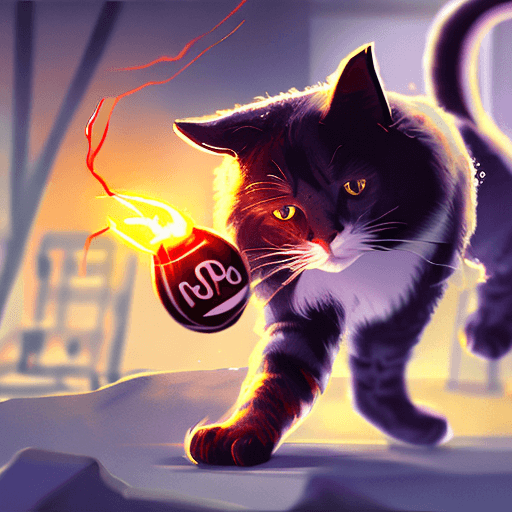} &
        \includegraphics[align=c, scale=0.16]{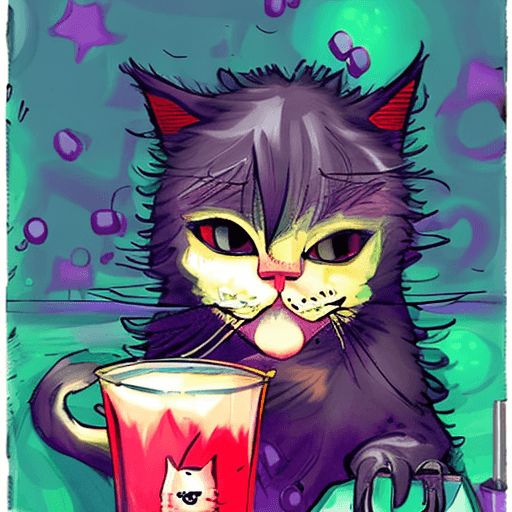}
    \\ [3pt]
        \begin{tabular}[c]{@{}c@{}}
            {\tiny the cat karakl drinks an energy} \\ [-5pt]
            {\tiny drink, concept art, wlop, digital} \\ [-5pt]
            {\tiny painting, trending on artstation,} \\ [-5pt]
            {\tiny highly detailed, epic composition,} \\ [-5pt]
            {\tiny official media, 8 k uhd}
        \end{tabular} &
        \begin{tabular}[c]{@{}c@{}}
            {\tiny th\^{}cat dryillustration} \\ [-5pt]
            {\tiny ilaypatreon atenefanart energy} \\ [-5pt]
            {\tiny drink drink overview digitalwiki} \\ [-5pt]
            {\tiny sergey igor rak kettcost} \\ [-5pt]
            {\tiny cg inna cg advise environment}
        \end{tabular} &
        \begin{tabular}[c]{@{}c@{}}
            {\tiny " cat energy drink illustration} \\ [-5pt]
            {\tiny ), archdmitpol ivan ks} \\ [-5pt]
            {\tiny cg \emojilight digitally visualization} \\ [-5pt]
            {\tiny deviantart patreon xiv}
        \end{tabular} &
        \begin{tabular}[c]{@{}c@{}}
            {\tiny fanart aneous art cat} \\ [-5pt]
            {\tiny patreon digitalcinematic} \\ [-5pt]
            {\tiny rendered energy drink} 
        \end{tabular} &
        \begin{tabular}[c]{@{}c@{}}
            {\tiny fanart cat drink}
        \end{tabular}
    \\
        \includegraphics[align=c, scale=0.16]{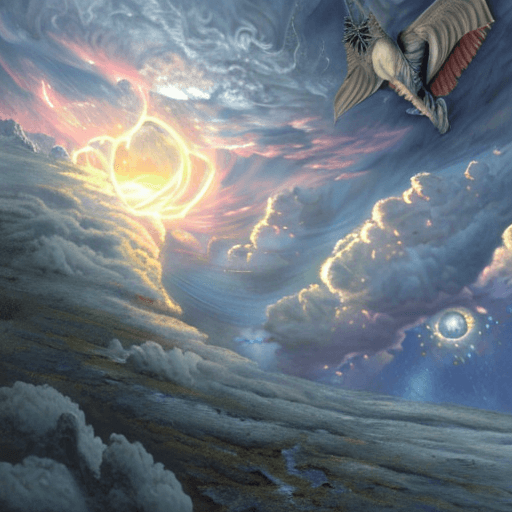} & 
        \includegraphics[align=c, scale=0.16]{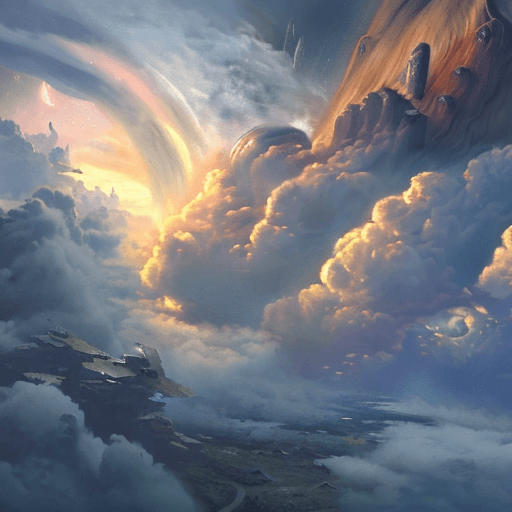} &
        \includegraphics[align=c, scale=0.16]{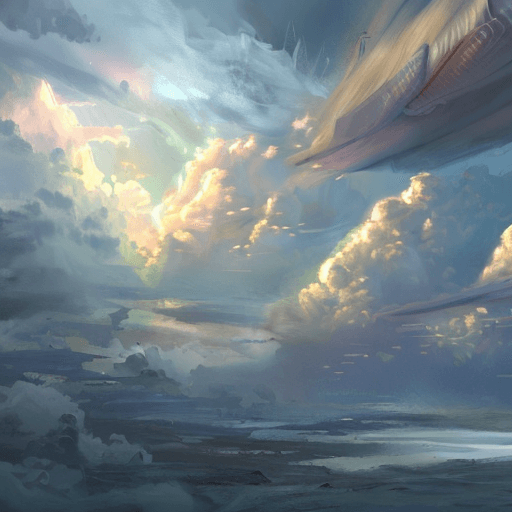} &
        \includegraphics[align=c, scale=0.16]{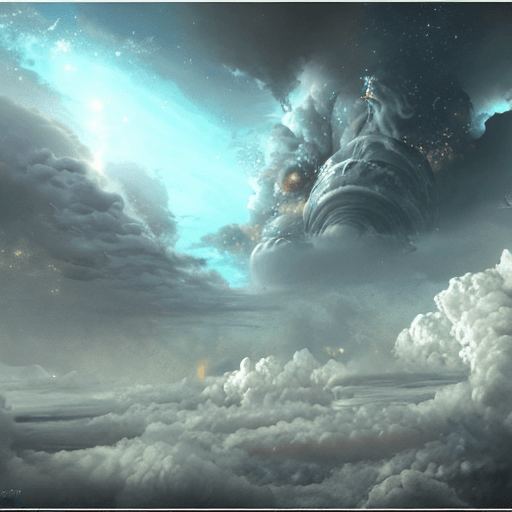} &
        \includegraphics[align=c, scale=0.16]{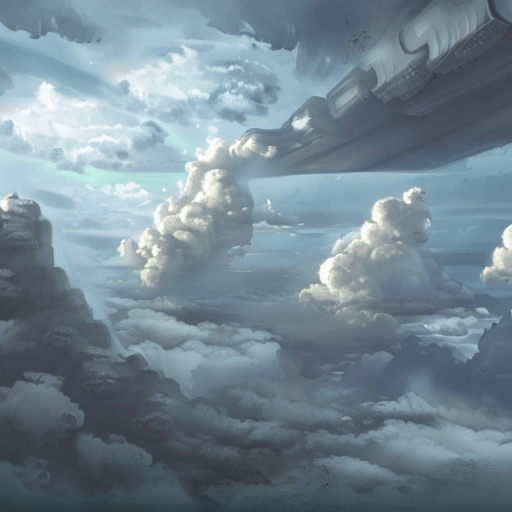}
    \\ [3pt]
        \begin{tabular}[c]{@{}c@{}}
            {\tiny Cloudscape by Adam Paquette,} \\ [-5pt]
            {\tiny nebula gasses in the background} \\ [-5pt]
            {\tiny by Gene Raz Von Edler,} \\ [-5pt]
            {\tiny fantasy magic angel concept} \\ [-5pt]
            {\tiny art from deviantart by Donato} \\ [-5pt]
            {\tiny Giancola, Rendered in Octane,} \\ [-5pt]
            {\tiny cinematic, Highly Detailed}
        \end{tabular} &
        \begin{tabular}[c]{@{}c@{}}
            {\tiny jesci vast clouds painting} \\ [-5pt]
            {\tiny cng fantasy biomedical fantasy} \\ [-5pt]
            {\tiny pulp hel picture nasa rpg} \\ [-5pt]
            {\tiny convergence patreon seuntotyotpo} \\ [-5pt]
            {\tiny mauricio acomzog lonler} \\ [-5pt]
            {\tiny ........ (© $<$ go}
        \end{tabular} &
        \begin{tabular}[c]{@{}c@{}}
            {\tiny clouds scenic scifi} \\ [-5pt]
            {\tiny maverbbhuttoillustration afm} \\ [-5pt]
            {\tiny criticalrolefanart conceptart} \\ [-5pt]
            {\tiny clouds ¯\textbackslash), sergey darrell} \\ [-5pt]
            {\tiny dewey royo faa} \\ [-5pt]
            {\tiny bild magelandscape}
        \end{tabular} &
        \begin{tabular}[c]{@{}c@{}}
            {\tiny return oung christensen fantasy} \\ [-5pt]
            {\tiny clouds skies colossus nebula} \\ [-5pt]
            {\tiny conceptart cinematic} \\ [-5pt]
            {\tiny rendering emporium}
        \end{tabular} &
        \begin{tabular}[c]{@{}c@{}}
            {\tiny scifi fantasy conceptart clouds}
        \end{tabular}
    \end{tabular}
    \caption{Prompt distillation. With fewer tokens, the hard prompts can still generate images very similar in concept to the original.}
    \label{fig:clip-prompt-distillation}
    \vspace{-.4cm}
\end{figure*}

\subsection{Style Transfer}
\looseness -1 The proposed approach can also be easily adapted to style transfer. We follow the setting investigated with soft prompts in \citet{gal2022image} but with our hard prompts. Given several examples that share the same style, we extract their shared style characteristics into a single hard prompt and use this prompt to apply the style to new objects or scenes. \cref{fig:clip-style} presents two examples of style transfer, showing that our method can easily embed the shared style elements in the prompt and apply them to novel concepts. Templates and learned prompts can be found in \cref{app:clip}.

\subsection{Prompt Concatenation}

Learned hard prompts are also very useful as composable building blocks for intricate scenes. We test this in \cref{fig:clip-merge}, where we separately generate prompts for two unrelated images, and then fuse both images by concatenating their prompts. We find that even different concepts, such as painted horses on a beach and a realistic sunset in a forest can be combined via their generated prompts.

\subsection{Prompt Distillation}
Another application where we can use our prompt optimization method is prompt distillation, reducing the length of prompts while preserving their capability. Distillation is useful in situations where the text encoder of the diffusion model has a limited maximum input length, such as the CLIP model, which has a maximum input length of $77$ tokens. Also, long prompts may contain redundant and unimportant information, especially when hand-crafted, so we aim to distill their essence, preserving only important information in the prompt.
We optimize a shorter prompt to match the features of the longer prompt simply based on its text encoder $f$. Given a target prompt's embedding $\mathbf{P_{\text{target}}}$ and learnable embedding $\mathbf{e}$, we simply modify our loss into: $\mathcal{L} = 1 - Sim(f(\mathbf{P_{\text{target}}}), f(\mathbf{P}))$. We define the distillation ratio by $|\mathbf{P}| / |\mathbf{P_{\text{target}}}|$.

In \cref{fig:clip-prompt-distillation}, we show images generated by the original prompts and the distilled prompts with four different distillation ratios: $0.7$, $0.5$, $0.3$, $0.1$. We see here that even with only $3$ or $4$ tokens, the hard prompts can still generate images very similar in concept to the original, successfully distilling the longer human-made instructions.

\vspace{-.2cm}
\section{Discrete Prompt Tuning with Language Models}\label{sec:lm-experiments}

\begin{table*}[t]
    \centering
    \small
    \caption{Accuracy ($\%$) and standard error on the SST-2 validation set across five prompts for each method learned on GPT-2 Large and transferred to larger models with 1.3B to 6.7B parameters. The baseline accuracy of a \textit{soft prompt} is $\mathbf{93.35_{\pm0.01}}$ (optimized for GPT-2 Large), but cannot be transferred across models.
    Empty$_{\text{Template}}$ refers to no prompt at the front but containing the predetermined template.} \label{tab:discrete_prompt_transferability}
    \begin{tabular}{cccccc}
    \toprule
 \multirow{2}{*}{Method} & GPT-2 Large & GPT-2 XL & T5-LM-XL & OPT & OPT \\
  & (755M, \textbf{Source}) & (1.3B) &  (3B) & (2.7B) &  (6.7B)\\ \toprule
Empty$_{\text{Template}}$ & 80.84 & 73.85 & 52.75 & 72.48 & 58.72 \\
AutoPrompt$_{\text{SGD}}$ & $87.56_{\pm0.35}$ & $78.19_{\pm2.68}$ & $56.01_{\pm1.67}$ & $73.69_{\pm1.63}$ & $65.28_{\pm1.75}$ \\
FluentPrompt & $\textbf{88.33}_{\pm0.35}$ & $78.53_{\pm2.82}$ & $55.64_{\pm0.59}$ & $70.39_{\pm2.08}$ & $61.74_{\pm1.25}$ \\
PEZ$_{\text{No Fluency}}$ (Ours) & $88.12_{\pm0.15}$ & $77.8_{\pm3.45}$ & $61.12_{\pm2.94}$ & $76.93_{\pm1.29}$ & $71.72_{\pm3.16}$ \\
PEZ$_{\text{Fluency}}$ (Ours) & $88.05_{\pm0.55}$ & $\textbf{79.72}_{\pm3.26}$ & $\textbf{63.30}_{\pm2.30}$ & $\textbf{77.18}_{\pm3.82}$ & $\textbf{72.39}_{\pm1.82}$ \\

\bottomrule
    \end{tabular}
\vspace{-.3cm}
\end{table*}

In the text-to-text setting, the goal of \cref{alg:pez} is to discover a discrete sequence of tokens, the hard prompt, that will prompt the language model to predict the outcome of a classification task.
Since an important property of text is its fluency, \citet{shi2022toward} find that fluency can increase a prompt's readability and performance. Thus, we define the optimization objective in this section as a weighted function of task loss and fluency loss,
$$\mathcal{L} = (1-\lambda_{\text{fluency}})\mathcal{L}_{\text{task}} + \lambda_{\text{fluency}}\mathcal{L}_{\text{fluency}}.$$
We set $\lambda=0.003$ similar to \citet{shi2022toward} for all methods, and we ablate our method without fluency ($\lambda = 0$), which we denote as \textit{no fluency}. 
We set out to show that hard prompts generated by this approach are successful both when transferring between a number of transformer-based language models, and also when used to discover prompts in few-shot settings. An attractive quality of these prompts, especially for language applications, is that they can be optimized on smaller language models and then transferred to other, much larger models.

\subsection{Datasets and Setup}
We evaluate \cref{alg:pez} against related algorithms on three classification tasks,
two sentiment analysis tasks, SST-2 \citep{socher-etal-2013-recursive} and Amazon Polarity \citep{10.1145/2507157.2507163}, and a 4-way classification task, AGNEWS \citep{zhang2015character}.
We build on the setting explored in \citet{ding-etal-2022-openprompt} and optimize hard prompts using GPT-2 Large (774M parameters) \citep{radford_language_2019} with the Adafactor optimizer \citep{shazeer2018adafactor} and a
batch size of 32 \citep{lester-etal-2021-power}. 
We provide details for prompt templates and verbalizers in \cref{tab:template_verbalizer}.


\looseness -1 \textbf{Transferability Set-up.}
To test transferability, we generate prompts from GPT-2 Large for 5000 steps.
We then select the five prompts with the highest average validation accuracy for each technique and test them on larger models. We test the transferred text on: GPT-2 XL, T5-LM-XL, OPT-2.7B, and OPT-6B \citep{radford_language_2019, lester_power_2021, zhang2022opt}, verifying the reliability of the proposed algorithm over related techniques and testing whether the hard prompt can reliably boost performance. Thus, we also consider a baseline of empty prompts, with only the template.

\textbf{Few-Shot Setup.}
For the few-shot setting, we optimize each prompt for 100 epochs
on GPT-2 Large on the AGNEWS dataset, where we sample two examples ($k=2$) and four examples ($k=4$) from each class to obtain the training set. Additionally, we create a holdout set of the same size, and finally validate the prompts on the entire validation set. 

\subsection{Results}
We verify that our method is comparable to other methods in the sentiment analysis setting and outperforms the other methods on AGNEWS by about $2\%$. See \cref{tab:discrete_prompt_tuning_text2text} for details. 

\looseness -1 \textbf{Prompt Transferability.}
\cref{tab:discrete_prompt_transferability} shows for each method the five prompts trained on GPT-2 Large transferred to other LLMs. Interestingly, simply scaling a model--with no additional training--does not guarantee that performance will scale accordingly.\footnote{A quick experiment with and without the template on GPT-2 Large and XL showed that the template boosts performance differently for different models.} We see that all gradient-based methods are able to transfer compared to evaluating just the template, finding that our prompts trained with the fluency constraint transfer better than the other prompts. 
Additionally, we can see the largest boost from OPT-6.7B with our fluent method with about a $14\%$ increase over just the template baseline. Additionally, we see our AGNEWS prompts are able to transfer from GPT-2 Large to GPT-2 XL in \cref{tab:transfer_agnews} of the Appendix. 

\begin{table}[t]
    \centering
    \small
   \caption{Average validation accuracy with standard error on AGNEWS with $k$ examples/shots per class using early stopping (including soft prompt) for all methods across 100 seeds for three tokens \textbf{append to the end of the text} similar to the original template (``It was about''). We set $\lambda=0.03$ for these experiments. ``Empty'' is the template with no additional prompt.}
    \label{tab:few_shot}
     \begin{tabular}[width=\columnwidth]{ccc} \toprule
Method  &   $k$=2   &   $k$=4\\ \midrule
Empty$_{\text{Template}}$   &       58.34     &         58.34\\ 
PEZ$_{\text{No Fluency}}$ (Ours)& $70.07_{\pm0.81}$    &   $73.99_{\pm0.45}$\\
PEZ$_{\text{Fluency}}$  (Ours) &   $70.93_{\pm0.60}$    &  $ 74.15_{\pm0.48}$\\
\midrule
Soft Prompt &   $74.92_{\pm0.58}$   &   $79.93_{\pm0.36}$\\ \bottomrule
    \end{tabular}
    \vspace{-.3cm}
\end{table}
\textbf{Prompt Discovery.}
 \cref{tab:few_shot} shows that even with just a few shots, we can achieve high validation accuracy compared to our prepended counterparts. It is worth noting that each few-shot run takes about $5$ minutes.
 
\looseness -1 We run $100$ seeds where the training set contains $k$ samples from each class and also qualitatively examine the top prompts. Although many of the prompts are non-interpretable, many are also coherent. For example, even for $k=2$, some of the prompts included news sources like ``\textit{BBC}'', while other prompts find new approaches to the news classification task considering the text coming from a blog: ``\textit{Brian blog,}'' or ``\textit{Blog Revolution analyze}.'' Due to the efficiency of these gradient-based methods, these methods can allow new ways for prompt engineers to discover novel prompts.

\section{Safety Concerns}
\begin{figure}
    \includegraphics[width=0.45\textwidth]{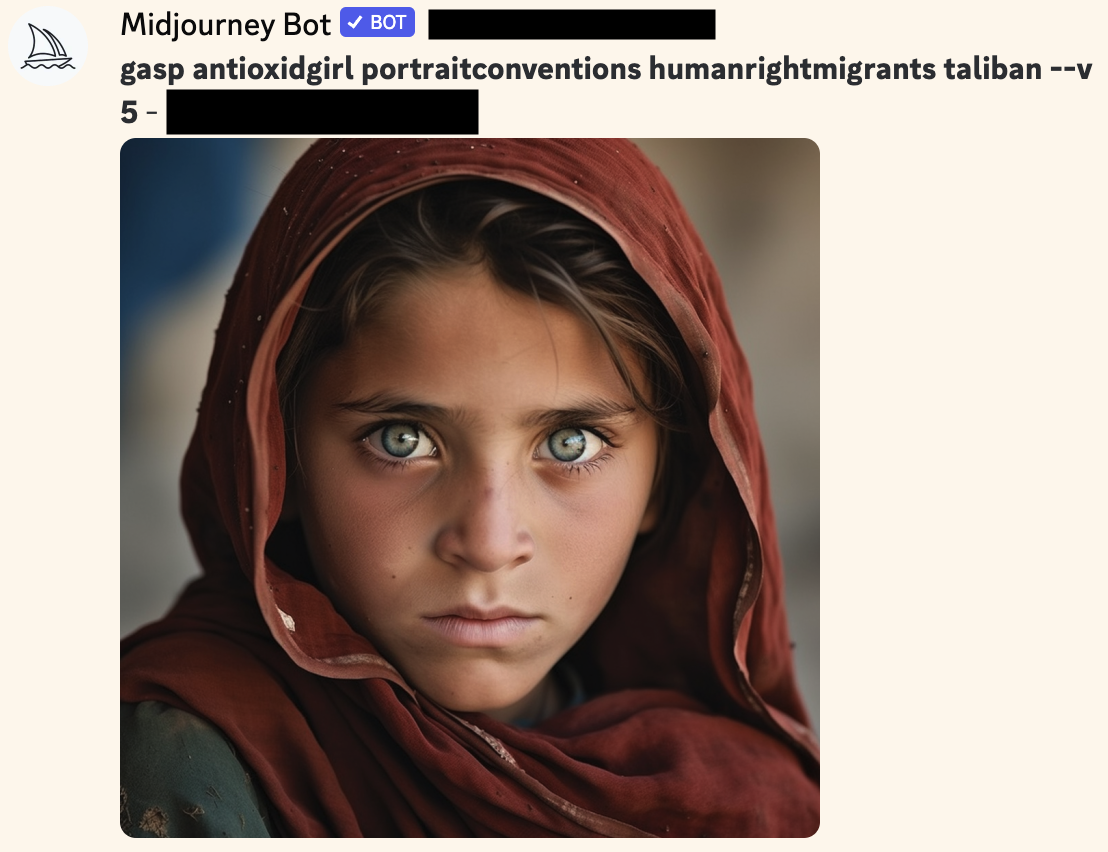}
    \caption{Generated copyrighted image via \texttt{Midjourney}. Here, requested from the API only for research purposes.}
    \label{fig:afgan_girl}
    \vspace{-.5cm}
\end{figure} 

Token or word-level content filters are often used in text-to-image diffusion model APIs to prevent the generation of NSFW or copyrighted content. For instance, the image generation API \texttt{Midjourney} has banned prompts containing the substring ``Afghan'' due to a copyright issue with the famous photo of an Afghan girl \footnote{\url{https://en.wikipedia.org/wiki/Afghan_Girl}}.

However, prompt optimization can be used as a mechanism to bypass simple rule-based content filters. PEZ can generate a prompt that avoids banned tokens, yet still matches textual features with the original target prompt ``Afghan girl.'' \cref{fig:afgan_girl} shows the output of \texttt{Midjourney} using an optimized prompt which successfully reproduces the banned image without containing the banned word ``Afghan.'' Note that the prompt seems to incorrectly associate the subject of the image, Sharbat Gula, with the Taliban.

Even if a defender now iterates the block-list and bans additional words from the adversarial prompt, an attacker can consistently optimize around addition content restrictions, as we show in supplementary material \cref{fig:afghan_girl_iterative}. Overall, we suspect that only complete feature-based content detectors have the potential to mitigate these concerns for model owners \citep{Rando2022RedTeamingTS}.

\section{Conclusion}
We propose a new method that utilizes continuous embeddings to reliably optimize hard prompts. The key advantage of our method is the use of continuous, i.e. soft, prompts as intermediate variables during the optimization of hard prompt tokens, leveraging gradient-based optimization. This way, the algorithm selects locations in embedding space where discrete embeddings are useful, rather than simply optimizing a soft prompt and later projecting onto nearby token embeddings in the hopes that these nearby hard prompts will perform well too.
Additionally, as our method utilizes gradients across all steps by accumulating them into the soft prompt, this process makes optimization more robust to learning rates and potential noise in the data.

Although our work makes progress toward prompt optimization, the community's understanding of language model embedding space is still in its infancy, and a deeper understanding of the geometry of the embedding space will likely enable even stronger prompt optimization in the future.

Overall, we show through our experiments that hard prompts can be easily generated and flexibly used in practical applications. Yet, a limitation of hard prompts is that even though they are human-readable, they may still contain several un-interpretable tokens.  Additionally, hard prompts may possibly extract harmful phrases or sensitive content from a language model's training data.  Even though we did not observe specific instances of this behavior, it is a concern that should be taken into account in future applications.

\section{Acknowledgements}
This work was made possible by the Office of Naval Research (N000142112557), the ONR MURI program, the National Science Foundation (IIS-2212182), and Capital One Bank.
\bibliography{refs, auto_refs}
\bibliographystyle{icml2023}

\appendix

\clearpage
\section{Appendix} \label{app:appendix}
\subsection{Additional Results for Prompt Inversion with CLIP}
\label{app:clip}
\begin{figure*}[t]
    \centering
    \begin{tabular}{c@{\hspace{20pt}}c@{\hspace{2pt}}c@{\hspace{2pt}}c@{\hspace{2pt}}c}
        Target Image & \multicolumn{4}{c}{Generated Image with Learned Hard Prompt} 
    \\
        \includegraphics[align=c, scale=0.16]{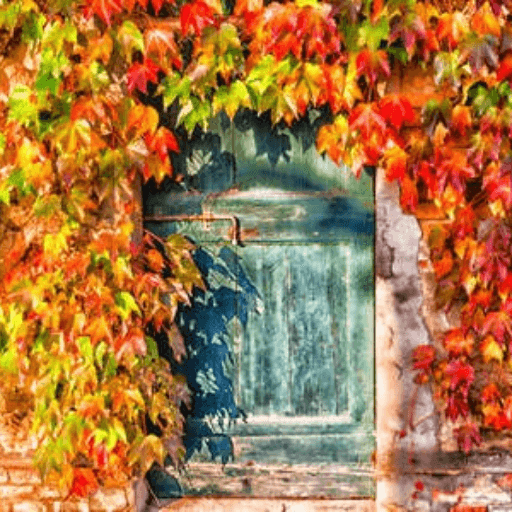} & 
        \includegraphics[align=c, scale=0.16]{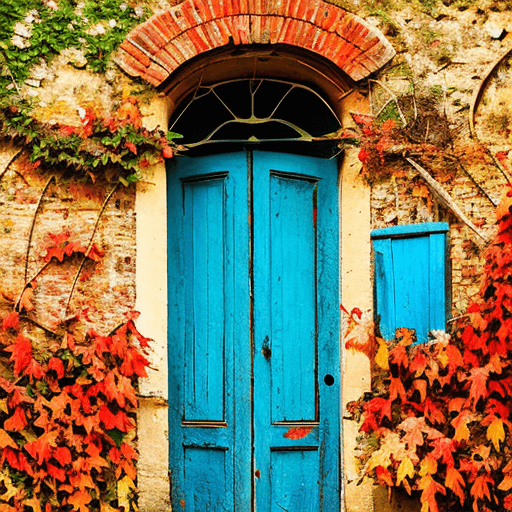} &
        \includegraphics[align=c, scale=0.16]{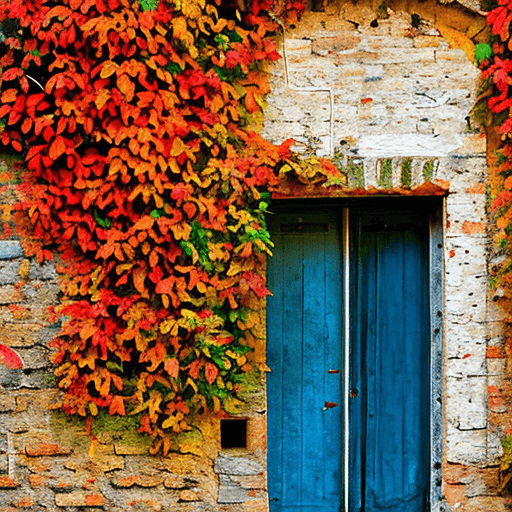} &
        \includegraphics[align=c, scale=0.16]{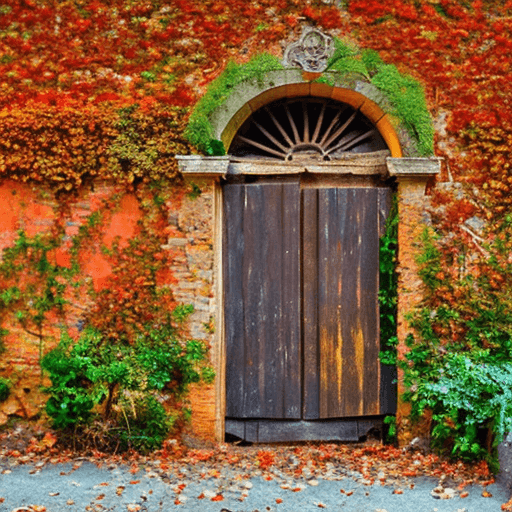} &
        \includegraphics[align=c, scale=0.16]{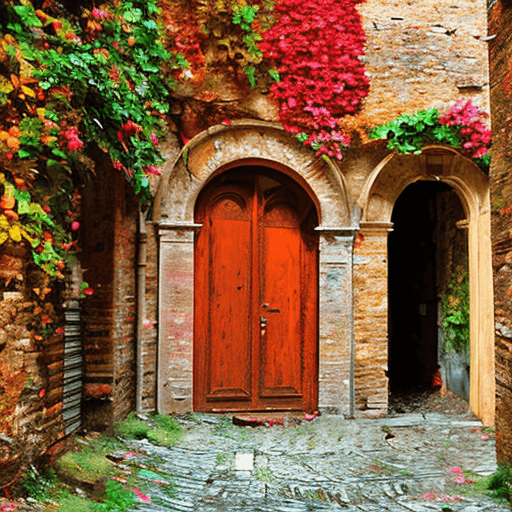}
    \\ [3pt]
        &
        \multicolumn{4}{c}{\begin{tabular}[c]{c} 
            {\scriptsize ohmydoor tuscany dickens ruin colorful fall d}
        \end{tabular}}
    \\
        \includegraphics[align=c, scale=0.16]{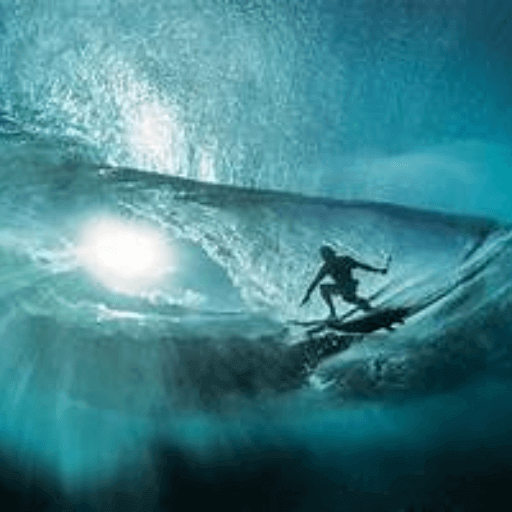} & 
        \includegraphics[align=c, scale=0.16]{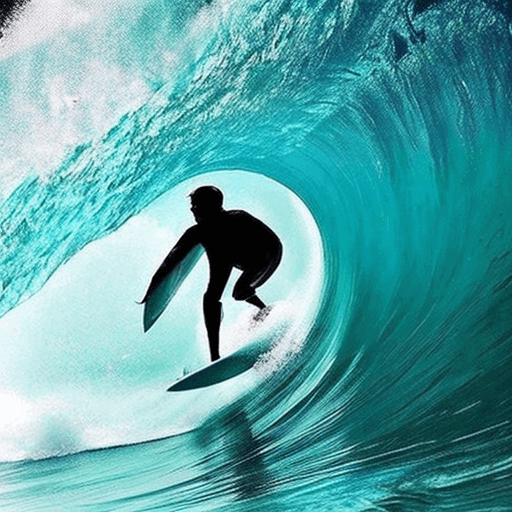} &
        \includegraphics[align=c, scale=0.16]{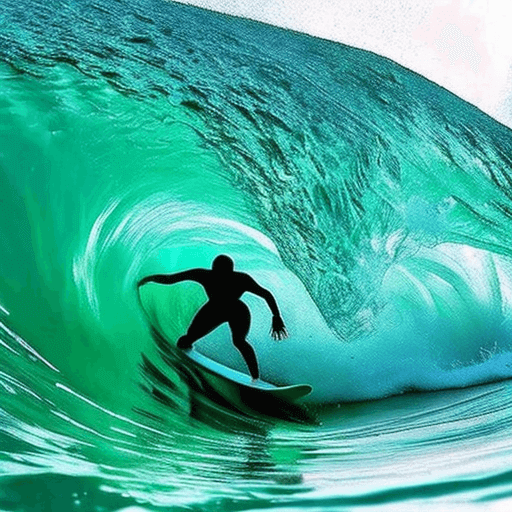} &
        \includegraphics[align=c, scale=0.16]{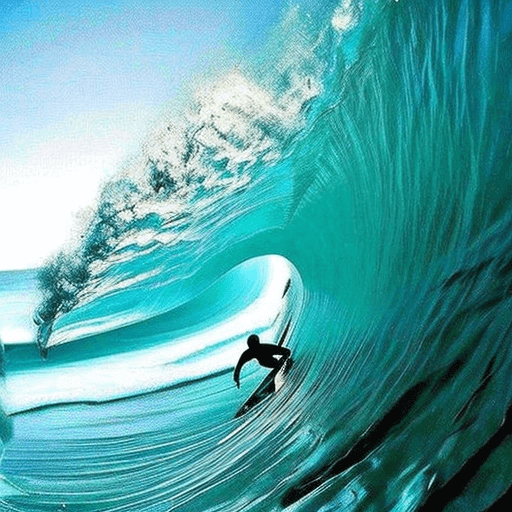} &
        \includegraphics[align=c, scale=0.16]{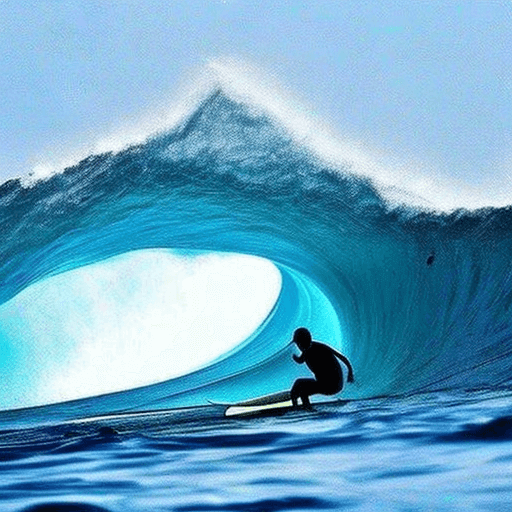}
    \\ [3pt]
        &
        \multicolumn{4}{c}{\begin{tabular}[c]{c} 
            {\scriptsize translucent abyss assaulted surfing featured regrann nbappinterest}
        \end{tabular}}
    \\
        \includegraphics[align=c, scale=0.16]{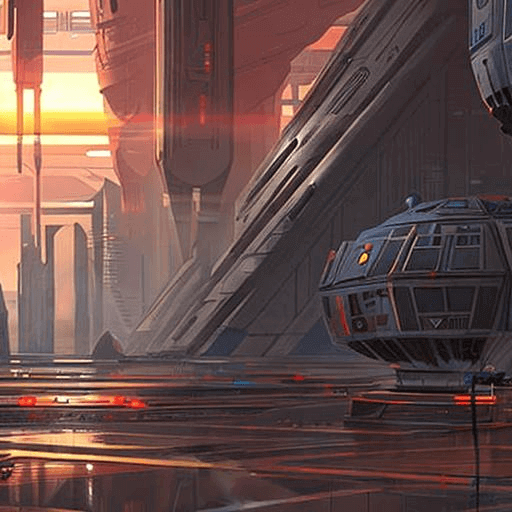} & 
        \includegraphics[align=c, scale=0.16]{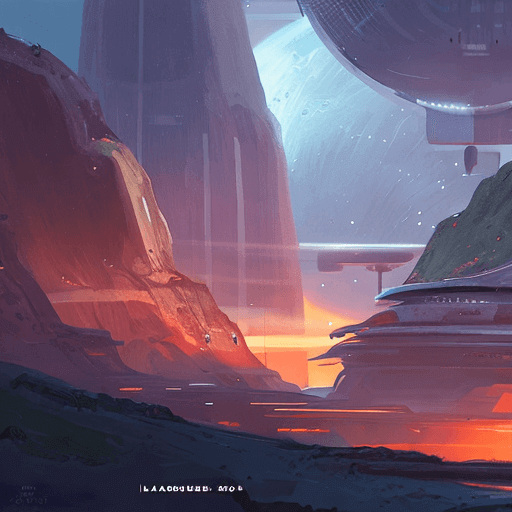} &
        \includegraphics[align=c, scale=0.16]{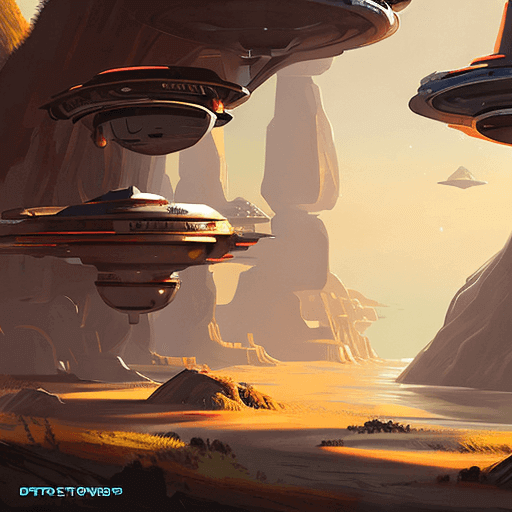} &
        \includegraphics[align=c, scale=0.16]{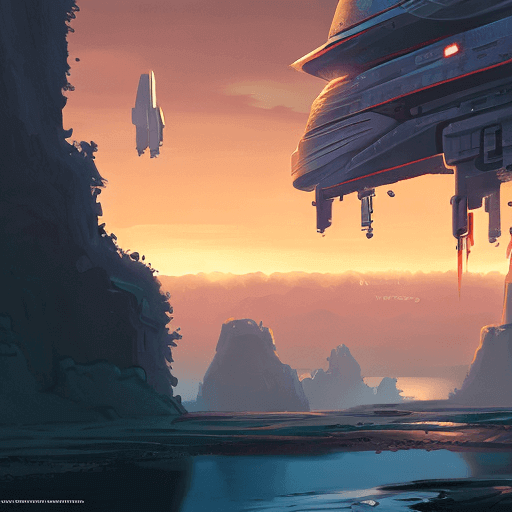} &
        \includegraphics[align=c, scale=0.16]{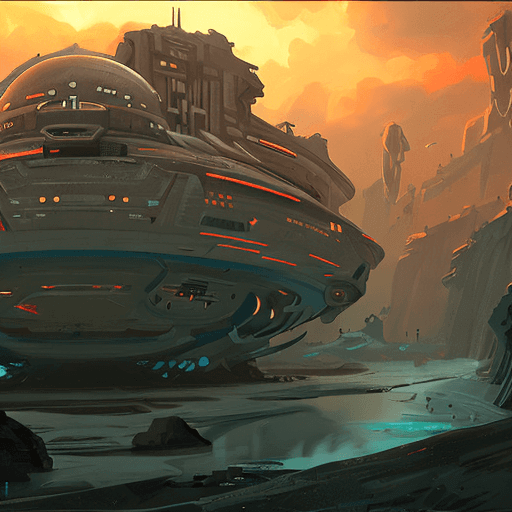}
    \\ [3pt]
        &
        \multicolumn{4}{c}{\begin{tabular}[c]{c} 
            {\scriptsize patreon alexandre dyk spaceship landscapes illustrtabletop painter}
        \end{tabular}}
    \\
        \includegraphics[align=c, scale=0.16]{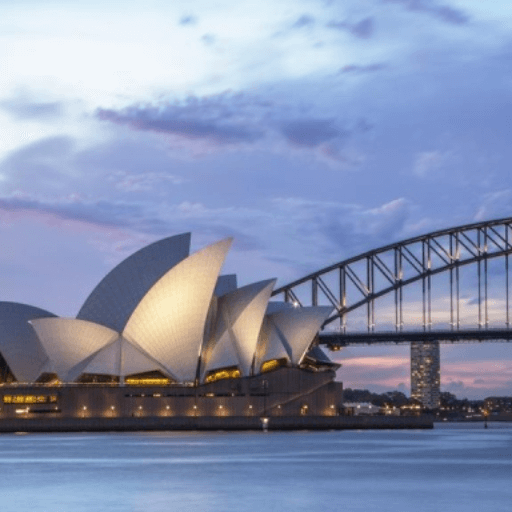} & 
        \includegraphics[align=c, scale=0.16]{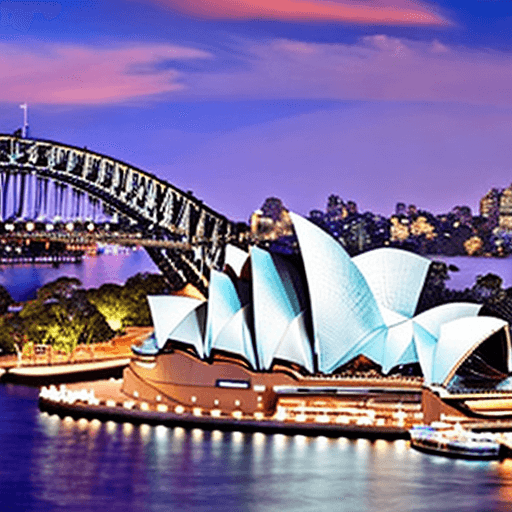} &
        \includegraphics[align=c, scale=0.16]{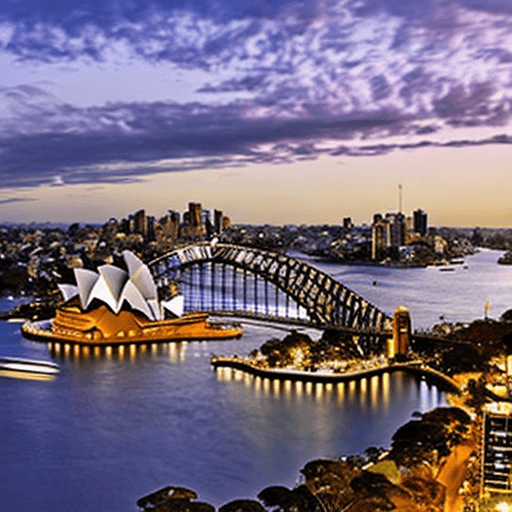} &
        \includegraphics[align=c, scale=0.16]{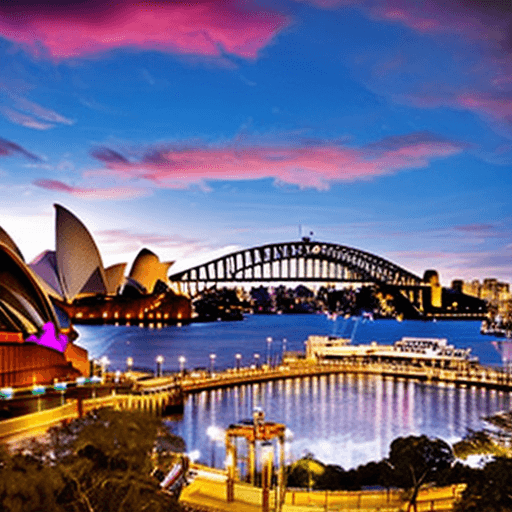} &
        \includegraphics[align=c, scale=0.16]{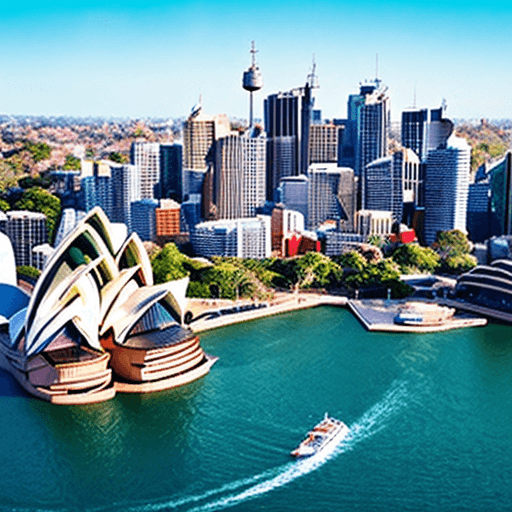}
    \\ [3pt]
        &
        \multicolumn{4}{c}{\begin{tabular}[c]{c} 
            {\scriptsize quiero amphitheatre launches sydney apac dua etf fed}
        \end{tabular}}
    \\
        \includegraphics[align=c, scale=0.16]{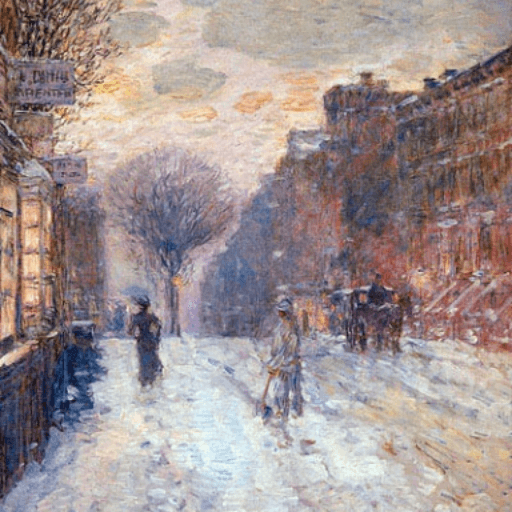} & 
        \includegraphics[align=c, scale=0.16]{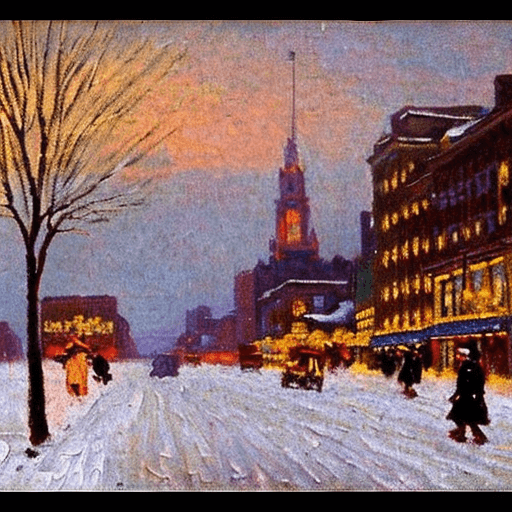} &
        \includegraphics[align=c, scale=0.16]{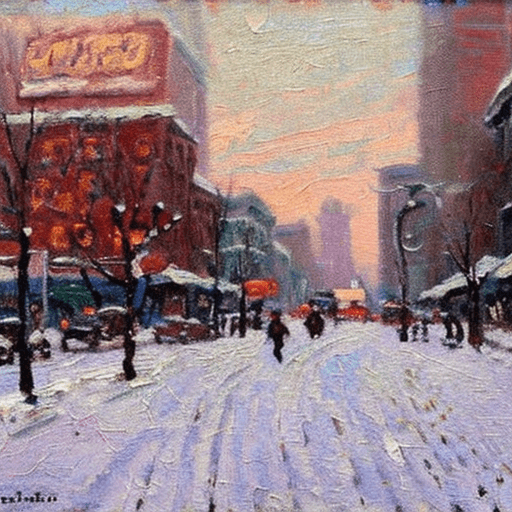} &
        \includegraphics[align=c, scale=0.16]{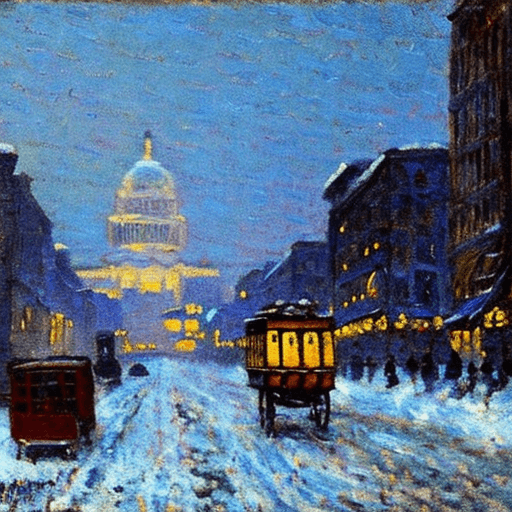} &
        \includegraphics[align=c, scale=0.16]{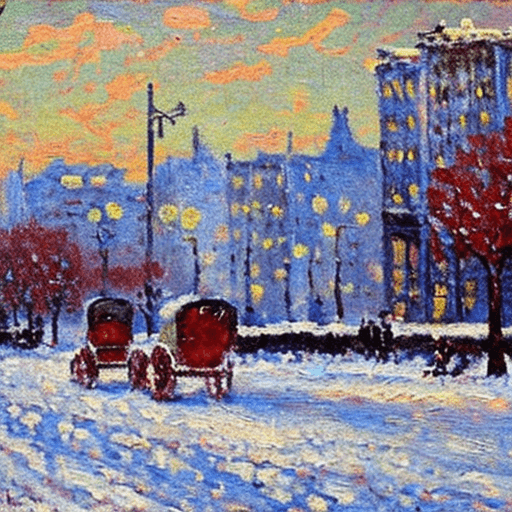}
    \\ [3pt]
        &
        \multicolumn{4}{c}{\begin{tabular}[c]{c} 
            {\scriptsize december montreal washington washingtonpopcorn impressionism paintings earliest}
        \end{tabular}}
    \\
        \includegraphics[align=c, scale=0.16]{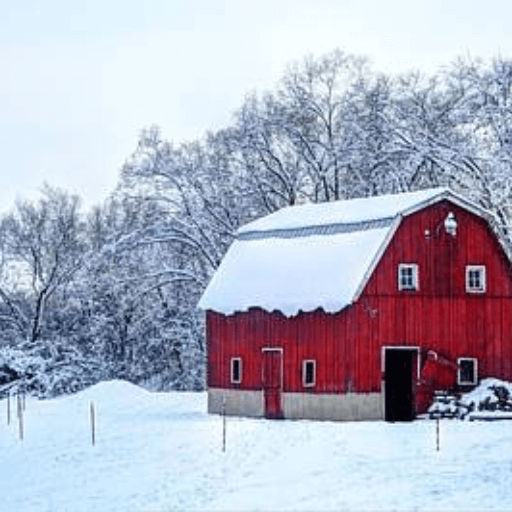} & 
        \includegraphics[align=c, scale=0.16]{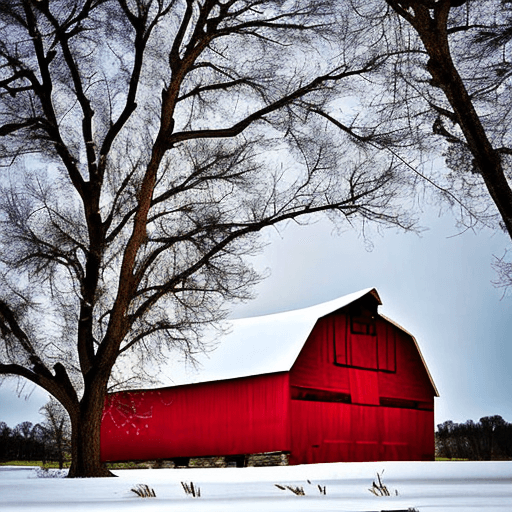} &
        \includegraphics[align=c, scale=0.16]{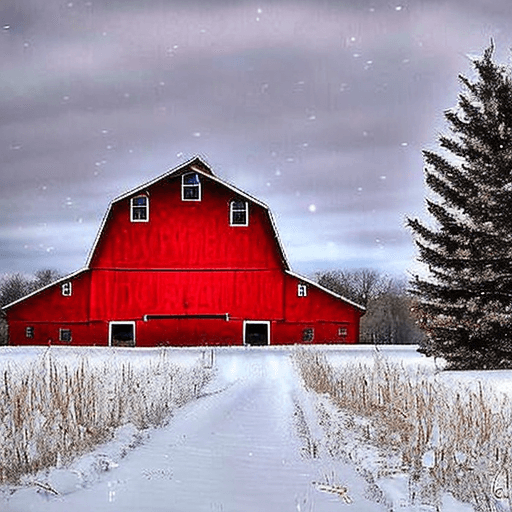} &
        \includegraphics[align=c, scale=0.16]{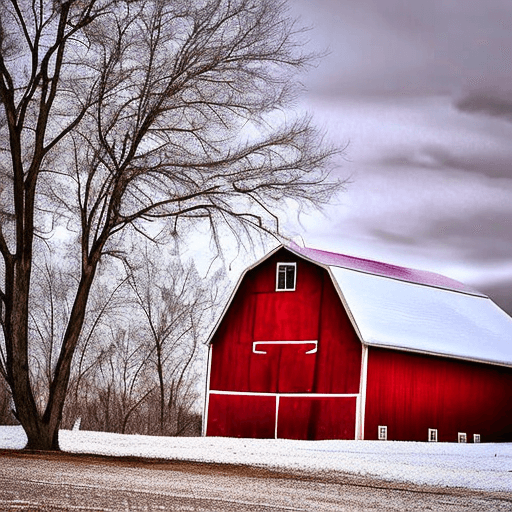} &
        \includegraphics[align=c, scale=0.16]{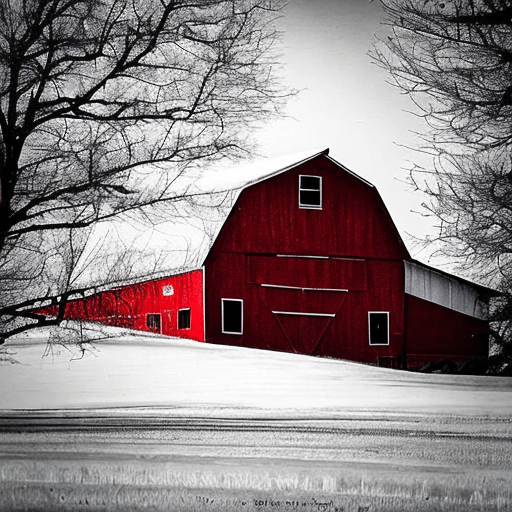}
    \\ [3pt]
        &
        \multicolumn{4}{c}{\begin{tabular}[c]{c}
            {\scriptsize wisconsin barn \emojipinkheart\emojipinkheart december by christy gendphotography}
        \end{tabular}}
    \end{tabular}
    \caption{Additional qualitative results with learned hard prompts.}
    \label{fig:clip-appendix}
\end{figure*}

We provide more qualitative results in \cref{fig:clip-appendix}.

For each example in \cref{fig:clip-style}, we use the following templates respectively: ``a tiger in the style of \{\}'', ``the streets of Paris in the style of \{\}'', ``a calculator in the style of \{\}'', ``a rocket in the style of \{\}'', where \{\} is replaced with the hard prompts:

\texttt{resonvillains stargazing illustration tutorials sma internationalwomensday watercolor fiberlilycamila yokohama -sorrow fluids latest }

\texttt{npr anime novels pureibanganesha irvin paints encapsulmondo illustrillustroversized sultanconan $\mbox{\textcent}$ }

for experiments 1 and 2, respectively.

\subsection{Additional Experiments and Details for Text-to-Text Hard Prompting}

\paragraph{Baseline Objective Formulations}
 Formally, we define a \textit{AutoPrompt$_\text{SGD}$} step as,
\begin{equation*} \label{eq:autoprompt}
    \mathbf{P}_{i+1} = \text{Proj}_{\mathbf{E}}[\mathbf{P}_i - \eta \nabla_{\mathbf{P}_i} \mathcal{L}(\mathcal{B}(\mathbf{P}_i, X_i), Y_i, \theta)]
\end{equation*}{}
Additionally, define \textit{FluentPrompt} updates follows,
\begin{equation*}
    \mathbf{P}_{i+1} = \text{Proj}_{\mathbf{E}}[\mathbf{P}_i - \eta \nabla_{\mathbf{P}_i} \mathcal{L}(\mathcal{B}(\mathbf{P}_i, X_i), Y_i, \theta) + \sqrt{2 \eta \beta_i}z]
\end{equation*}

\paragraph{Details for \cref{sec:lm-experiments}}
For \cref{tab:discrete_prompt_tuning_text2text}, we report the best validation accuracy across three learning rates (0.1, 0.3, and 0.5), and for \textit{FluentPrompt} and \textit{AutoPrompt}$_{\text{SGD}}$ we used the learning reported (1, 3, and 10) and follow \citet{shi2022toward} for the remaining hyperparameters for \textit{FluentPrompt}. For these experiments, we \textit{prepend} our 10 token prompt to each input text. We employ early stopping for all methods using a hold-out set of 5000 examples for each dataset, evaluating every 100 steps. 

\cref{tab:discrete_prompt_tuning_text2text} shows that we are comparable to other methods in sentiment analysis and outperform the other methods on AGNEWS by about $2\%$. Examining the prompts, we find prompts are not coherent English for any of the methods. However, it does produce relevant tokens and phrases. For example, our method for SST-2 with the fluency constraint produced ``\textit{negative vibeThis immatureollywood MandarinollywoodThis energetic screenplay}.'' \footnote{Although we initialize the tokens with the label tokens, when examining the prompt over the optimization process, all tokens moved away from the initial tokens. This suggests that the process was able to relearn the class label.} This suggests the optimization process is finding relevant words to the task but lacks the ability to create full sentences.
\begin{table}[h]
 \vspace{-.5cm}
    \centering
    \small
    \caption{The template and verbalizer used for each dataset.}
    \begin{tabular}{ccc}
    \toprule
Dataset & Template                                                            & Verbalizer                                                                       \\ \toprule
SST-2   & \textless{}s\textgreater It was \textless{}mask\textgreater{}       & positive, negative                                                               \\ \midrule
Amazon  & \textless{}s\textgreater  It was \textless{}mask\textgreater{}       & positive, negative                                                               \\ \midrule
AGNEWS  & \textless{}s\textgreater It was about \textless{}mask\textgreater{} & \begin{tabular}[c]{@{}l@{}}politics, sports,\\ business, technology \end{tabular} \\ \bottomrule
    \end{tabular}

    \label{tab:template_verbalizer}
    \vspace{-.2cm}
\end{table}

\begin{table}[h]
\vspace{-.5cm}
    \centering
    \small
    \caption{Validation accuracy for 10 discrete tokens trained \textbf{prepended at the beginning of the input text}. Best accuracy across three learning with standard error reported over 5 speeds.}
    \begin{tabular}{cccc}
    \toprule
Method &SST-2&AGNEWS&Amazon\\ \midrule
AutoPrompt$_{\text{SGD}}$&$87.56_{\pm 0.35}$&$74.36_{\pm 0.47}$&$87.75_{\pm 0.17}$\\
FluentPrompt&$88.33_{\pm 0.35}$&$74.62_{\pm0.24}$&$87.42_{\pm0.18}$\\
PEZ$_{\text{No Fluency}}$(Ours) & $88.12_{\pm 0.15}$&$77.06_{\pm 0.20}$&$87.70_{\pm 0.21}$\\
PEZ$_{\text{Fluency}}$(Ours) & $88.05_{\pm 0.55}$&$76.94_{\pm 0.48}$&$87.78_{\pm 0.19}$\\ \midrule
Soft Prompt & $93.35_{\pm0.01}$ & $92.76_{\pm0.01}$ &$94.65_{\pm 0.01}$\\ \bottomrule
    \end{tabular}
    \label{tab:discrete_prompt_tuning_text2text}
    \vspace{-.2cm}
\end{table}

\begin{figure}[h]
    \centering
    \includegraphics[width=.8\columnwidth]{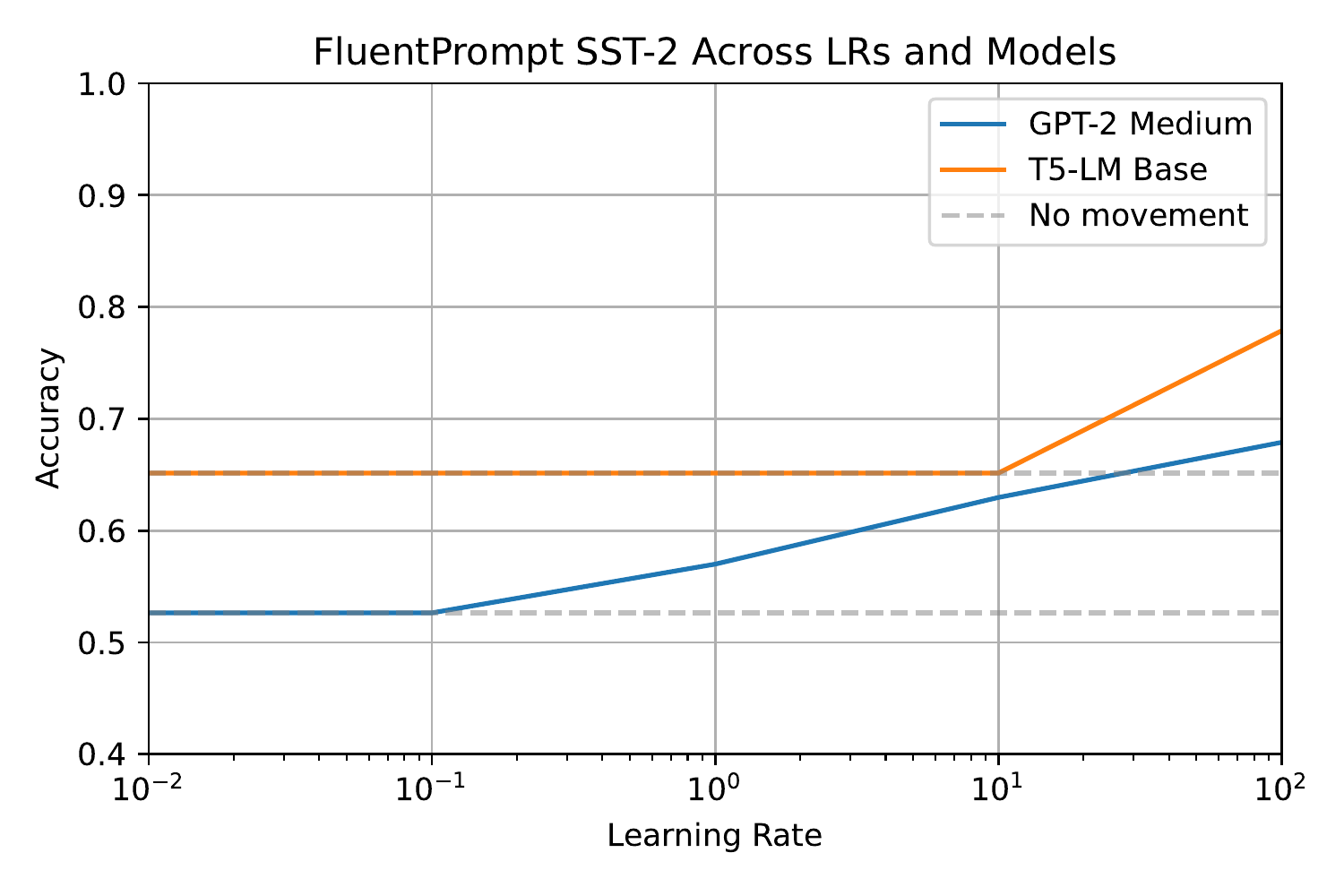}
    \caption{Displays that by projecting every step \textit{FluentPrompt}, and by extension \textit{AutoPrompt$_{\text{SGD}}$}, can be subject to some interesting learning rates that are very model dependent.}
    \label{fig:FluentPrompt_LRsweep}
\end{figure}

\begin{table}[h!]
  \centering
  \small
\caption{Shows the validation accuracy with standard deviation from transferring hard prompts learned on GPT-2 Large to GPT-2 XL.}\label{tab:transfer_agnews}
\begin{tabular}{ccc}
\toprule
Method                    & GPT-2 Large (755M) & GPT-2 XL (1.3B)   \\ \midrule
Empty$_{\text{template}}$ & {\color{gray} 58.34}              & 52.42            \\
AutoPrompt$_{\text{SGD}}$                & {\color{gray} $74.36_{\pm 0.47}$} & $63.79_{\pm3.61}$ \\
FluentPrompt              & {\color{gray} $74.62_{\pm0.24}$}  & $61.57_{\pm5.1}$  \\
PEZ$_{\text{No Fluency}}$(Ours)       & {\color{gray} $77.06_{\pm 0.20}$} & $59.45_{\pm8.63}$ \\
PEZ$_{\text{Fluency}}$(Ours)       & {\color{gray} $76.94_{\pm 0.48}$} & $67.59_{\pm2.67}$ \\
\bottomrule
\end{tabular}
\vspace{-.2cm}
\end{table}

\begin{figure*}[!h]
    \centering
    \includegraphics[width=1\textwidth]{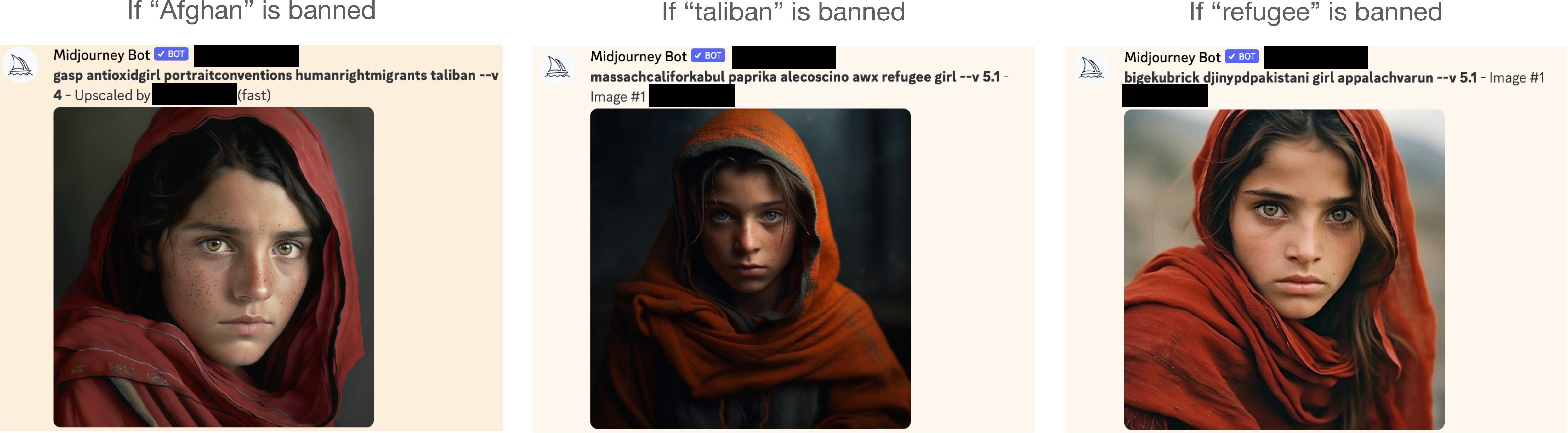}
    \caption{Iteratively evade \texttt{Midjourney} content filter and remove sensitive words/tokens.}
    \label{fig:afghan_girl_iterative}
\end{figure*}

\clearpage

\begin{table*}[!htb]
\centering
\caption{Quantitative results on learned hard prompts. We report the CLIP score between the original images and the images generated by the hard prompts}
\label{table:app-clip-main}
\begin{tabular}{ccccccc}
\toprule
                 & \#Tokens   & Requirement         & LAION & MS COCO & Celeb-A & Lexica.art \\ \midrule
AutoPrompt$_{\text{SGD}}$ & $8$ & CLIP                & $0.689$ & $0.669$   & $0.595$   & $0.702$   \\
FluentPrompt     & $8$  & CLIP                & $0.688$ & $0.671$   & $0.583$   & $0.702$      \\
PEZ (Ours)             & $8$   & CLIP                & $0.697$ & $0.674$   & $0.602$   & $0.711$      \\
CLIP Interrogator & $\sim77$  & CLIP + Bank + BLIP & $0.707$ & $0.690$   & $0.558$   & $0.762$      \\ \midrule
CLIP Interrogator without BLIP  & $\sim77$ & CLIP + Bank        & $0.677$ & $0.674$   & $0.572$   & $0.737$      \\
PEZ (Ours) + Bank      & $8$ & CLIP + Bank        & $0.702$ & $0.689$   & $0.629$   & $0.740$      \\ \midrule
CLIP Interrogator    & $8$ & CLIP + Bank + BLIP & $0.539$ & $0.575$   & $0.360$   & $0.532$      \\
CLIP Interrogator   & $16$ & CLIP + Bank + BLIP & $0.650$ & $0.650$    & $0.491$   & $0.671$      \\
CLIP Interrogator  & $32$ & CLIP + Bank + BLIP & $0.694$ & $0.663$   & $0.540$  & $0.730$      \\ \midrule
Soft Prompt   & $8$ & CLIP & $0.408$ & $0.420$   & $0.451$   & $0.554$      \\ \bottomrule
\end{tabular}
\end{table*}

\begin{figure*}
    \centering
    \includegraphics[width=0.45\textwidth]{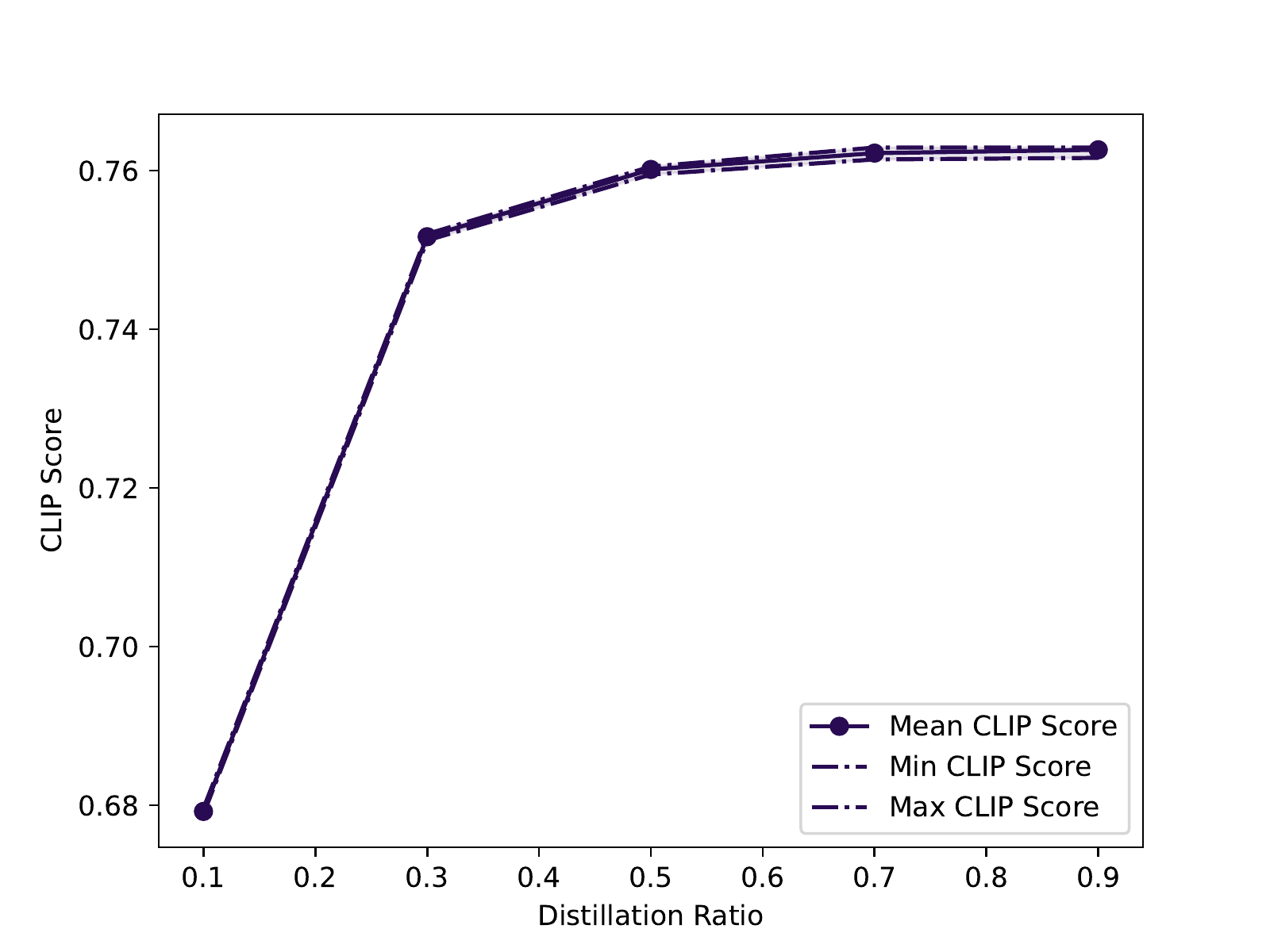}
    \caption{Quantitative results on prompt distillation with different distillation ratios. The CLIP score is calculated between the images generated by the original prompt and the images generated by the distilled prompt.}
    \label{fig:prompt-dist-plot}
\end{figure*}

\begin{figure*}
    \begin{center} 
    \includegraphics[width=\columnwidth]{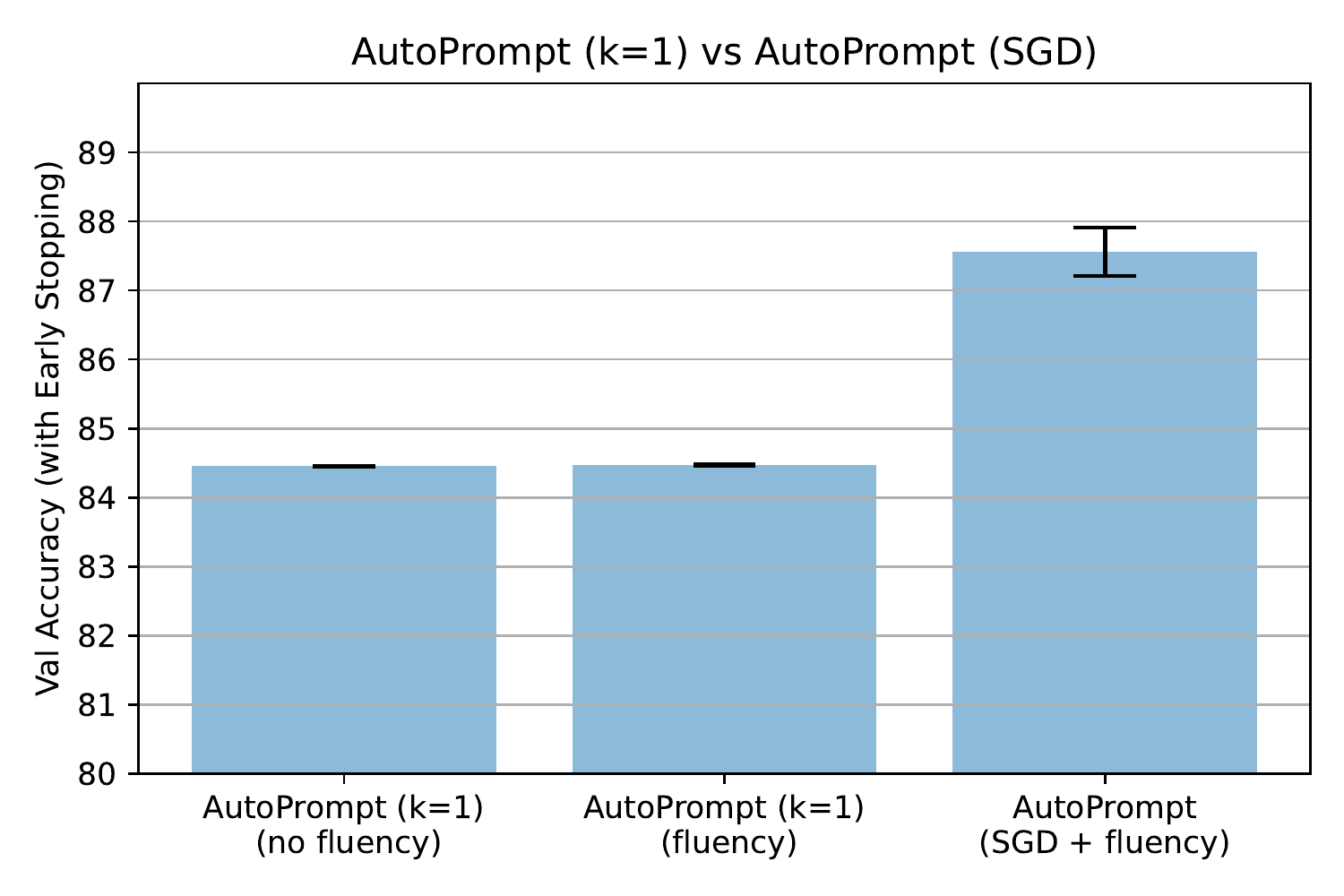}
    \caption{SST-2 validation accuracy comparing AutoPrompt (k=1) and AutoPrompt with SGD. From the figure, we can see that AutoPrompt SGD is better than AutoPrompt (k=1), where k is the number of candidates evaluated by the greedy process.}
    \label{fig:AutoPromptSGDvsK1}
    \end{center}
\end{figure*}

\end{document}